\RequirePackage{fix-cm}
\documentclass[twocolumn]{svjour3}          
\smartqed  
\usepackage{type1cm}        
\usepackage{makeidx}         
\usepackage{graphicx}        
\usepackage[bottom]{footmisc}

\usepackage[utf8]{inputenc} 

\usepackage{newtxtext}       %
\usepackage{newtxmath}       
\usepackage{hyperref}

\usepackage{siunitx}

\hypersetup{
   colorlinks,%
   citecolor=blue,%
   filecolor=black,%
   linkcolor=blue,%
   urlcolor=blue
}

\makeindex             

\usepackage{todonotes}

\begin{document}

\title{The Tracking Machine Learning challenge : Throughput phase
}

\titlerunning{TrackML : Throughput phase}        


\author{Sabrina Amrouche \and  
Laurent Basara \and
Paolo Calafiura \and
Dmitry Emeliyanov \and
Victor Estrade \and
Steven Farrell \and
C\'ecile Germain \and
Vladimir Vava Gligorov \and
Tobias Golling \and
Sergey Gorbunov \and
Heather Gray \and
Isabelle Guyon \and
Mikhail Hushchyn \and
Vincenzo Innocente \and
Moritz Kiehn \and
Marcel Kunze \and
Edward Moyse \and
David Rousseau \and
Andreas Salzburger \and
Andrey Ustyuzhanin \and
Jean-Roch Vlimant
}


\institute{
\footnote{participant} participant \\ 
Paolo Calafiura \email{pcalafiura@lbl.gov} \and Steven Farrell \email{sfarrell@lbl.gov} \and  Heather Gray \email{heather.gray@berkeley.edu} \at University of California, Berkeley CA, USA  and Physics Division, Lawrence Berkeley National Laboratory 
\and  Jean-Roch Vlimant \email{jvlimant@caltech.edu} \at California Institute of Technology, Pasadena CA, USA
\and  Vincenzo Innocente \email{vincenzo.innocente@gmail.com} \and  Andreas Salzburger \email{asalzburger@gmail.com} \at CERN, Geneva, Switzerland
\and  $^*$Dmitry Emeliyanov \email{d.emeliyanov@outlook.com} \at Particle Physics Department, Rutherford Appleton Laboratory, Didcot, United Kingdom
\and  $^*$Sergey Gorbunov \email{sergey.gorbunov@fias.uni-frankfurt.de} \at Goethe University Frankfurt, Germany 
\and  $^*$Marcel Kunze \email{Marcel.Kunze@uni-heidelberg.de} \at  Heidelberg University, Germany 
\and  Sabrina Amrouche \email{sabrina.amrouche@cern.ch} \and   Tobias Golling \email{Tobias.Golling@unige.ch} \and  Moritz Kiehn \email{Moritz.Kiehn@cern.ch} \at D\'epartement de Physique Nucl\'eaire et Corpusculaire, Universit\'e de Gen\`eve, Gen\`eve, Switzerland
\and  Edward Moyse \email{Edward.Moyse@cern.ch} \at Department of Physics, University of Massachusetts, Amherst MA, USA 
\and  Laurent Basara \email{laurent.basara@inria.fr},  Victor Estrade \email{estrade@lri.fr} \and  C\'ecile Germain \email{cecile.germain@inria.fr} \at LRI/TAU, Univ. Paris-Sud/INRIA/CNRS, Universit\'{e} Paris-Saclay, Gif-sur-Yvette, France
\and   Isabelle Guyon \email{guyon@chalearn.org} \at UPSud/INRIA Universit\'e Paris-Saclay, Orsay, France, and ChaLearn, Berkeley, CA, USA
\and contact: David Rousseau \email{rousseau@lal.in2p3.fr}, 
\and  Yetkin Yilmaz \email{yetkinyilmaz@gmail.com} \at Universit\'e Paris-Saclay, CNRS/IN2P3, IJCLab, 91405 Orsay, France 
\and  Mikhail Hushchyn \email{hushchyn.mikhail@gmail.com}
\and  Andrey Ustyuzhanin \email{andrey.u@gmail.com} \at National Research University Higher School of Economics and Yandex School of Data Analysis, Moscow, Russia
\and  Vladimir Vava Gligorov \email{Vladimir.Gligorov@cern.ch} \at LPNHE, Sorbonne Universit{\'e}, Paris Diderot Sorbonne Paris Cit{\'e}, CNRS/IN2P3, Paris, France
}


\date{Received: date / Accepted: date}

\maketitle

\abstract{

This paper reports on the second ``Throughput'' phase  of the Tracking Machine Learning (TrackML) challenge on the Codalab platform. 
As in the first ``Accuracy'' phase, the participants had to solve a difficult
experimental problem linked to tracking accurately the trajectory of particles as e.g. created at the Large Hadron Collider (LHC): given $O(10^5)$ points, the participants had to connect them into $O(10^4)$ individual groups that represent the particle trajectories which are approximated helical.
While in the first phase only the accuracy mattered, the goal of this second phase was a compromise between the accuracy and the speed of inference.
Both were measured on the Codalab platform where the participants had to upload their software. The best three participants had solutions with good accuracy and speed an order of magnitude faster than the state of the art when the challenge was designed. Although the core algorithms were less diverse than in the first phase, a diversity of techniques have been used and are described in this paper. The performance of the algorithms are analysed in depth and lessons derived.
}

\tableofcontents

\section{Introduction}

\label {s:intro}
The Tracking Machine Learning (TrackML) challenge took place in two phases, an Accuracy phase\cite{TrackMLAccuracy2019} in 2018 on the Kaggle platform\footnote{https://www.kaggle.com}, and a Throughput phase in 2018-2019 on Codalab\footnote{https://competitions.codalab.org}, preceded by a limited scope 2D prototype competition \cite{TrackMLRamp2017}. This paper is documenting in details the Throughput phase, which combined accuracy and inference speed, while only the minimal summary of the Accuracy phase is given (see \cite{TrackMLAccuracy2019} for details). The goal of these competitions was to reach out to a wider community to stimulate various approaches to tracking on a uniform set up,  as it has been done in the past on a variety of HEP issues~\cite{Rousseau:2020rnz}.

The LHC is a unique particle accelerator complex colliding protons at unprecedented energies. It allowed the Higgs boson discovery\cite{Aad:2012tfa,Chatrchyan:2012ufa} in 2012 as acknowledged by the 2013 Nobel prize in physics. It will collect data of increasing complexity and at an increasing rate with a large upgrade the so-called High Luminosity LHC (HL-LHC)\cite{ApollinariG.:2017ojx}  currently planned for 2027. The analysis pipelines of the proton collisions (or {\it events}) rely as an important step on the reconstruction of the trajectories of the particles within the innermost parts of the detector. 
The time to reconstruct the trajectories --- in a constant magnetic field these would follow a helical path --- from the measurements (3D points) is expected to increase faster than the projected computing resources. New approaches to pattern recognition are thus necessary to exploit fully the discovery potential of the HL-LHC. 
A typical event of this challenge would have about 100.000 points to be associated into about 10.000 trajectories. The state of the art was of order 10s per event on single CPU core when the challenge was designed\cite{ATLASfasttracking,CMSresources}. Given that 10 to 100 billion such collisions need to be processed each year, the importance of the increase of the reconstruction throughput becomes evident.

A dataset consisting of a relatively detailed simulation\cite{Gessinger:2020nne} of an LHC-like experiment has been created, listing for each event the measured 3D points, and the list of 3D points associated to a true track.
The dataset is large enough to allow for the training of data-intensive Machine Learning methods; the order of magnitude estimates are: ten thousand events, one billion measurement points, one hundred million trajectories (``tracks'') to be found.
In practice, the task is to build the list of 3D points belonging to each track. As usual, the solutions proposed by the participants were evaluated on a test set stripped of the ground truth. 
The final step of track reconstruction, i.e. the inference of the particle properties (track parameters) at the particle's origin, was not a goal of the challenge, given that the estimation of track parameters by applying fitting or other inference techniques is believed to be well understood and does not significantly drive the computing requirements.

For the Accuracy phase, participants had to upload a solution file (in csv format) indicating how the points are clustered (like for a typical competition on the Kaggle platform), while for the Throughput phase participants had to upload their software to the Codalab platform, on which it was executed in a controlled environment. By doing so, the resource usage was measured in a standardized way, and the Throughput score was then derived from a combination of the accuracy and the inference speed.

This paper is organised as follows. Section~\ref{sec:setup} details the setup of the competition, the changes to the dataset with regards to the Accuracy phase and accuracy score evaluation, the score and the details of the implementation on the Codalab platform. Section~\ref{sec:competition} narrates the competition as it happened. Section~\ref{sec:performance} details the performance of the algorithms submitted. The top three algorithms are then  detailed each in sections ~\ref{sec:gorbunov}, ~\ref{sec:emeliyanov} and ~\ref{sec:kunze}, respectively, and~\ref{sec:conclusion} is the Conclusion.

\section {Throughput phase setup}
\label{sec:setup}

This section details the setup of the competition, building on the Accuracy competition description in \cite{TrackMLAccuracy2019}.

\subsection {Dataset update}

\begin{figure}[ht]
\centering
\includegraphics[width=0.48\textwidth]{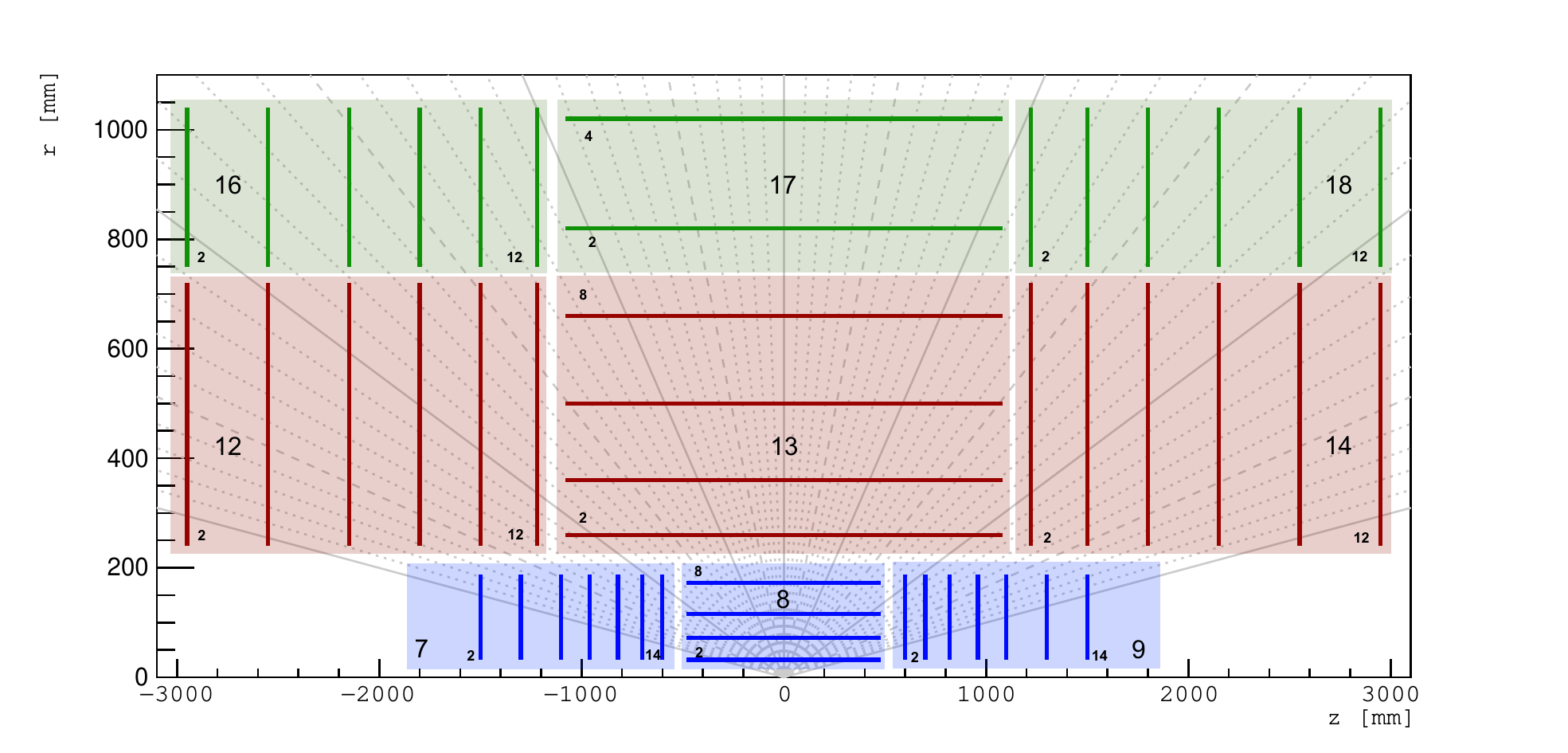}
\caption{Sketch of the TrackML detector as used in both the "Accuracy" and "Throughput" phase. Vertical lines indicate disks while horizontal lines indicate cylinders, all with the $z$ axis as axis of revolution. Three different sub detectors build the overall detector setup: a central pixel system (blue), enclosed by first a short strip (red) and then a long strip detector (green).}
\label{fig:DetectorTrackML}
\end{figure}

The dataset~\cite{salzburger_andreas_2018_4730157} for this Throughput phase is slightly different from the one \cite{salzburger_andreas_2018_4730167} for the Accuracy phase. It was produced with the fast detector simulation that is part of the ACTS project\cite{Gessinger:2020nne}. The detector setup, as pictured in Fig.~\ref{fig:DetectorTrackML}, was unchanged with respect to the Accuracy phase, and mimics a typical LHC general purpose experiment. The overall detector setup is as follows:
A central silicon pixel detector with the \SI{50}{\micro\metre}  square pixels
is enclosed by a Silicon short strip detector and an outermost long strip detector and embedded in a solenoidal magnetic field with a central field strength of 2~T.

Minor adaptions to the dataset have been made for the Throughput phase, predominantly to correct issues that have been identified with the Accuracy dataset (those issues, however, were checked to not have any impact on the outcome of the Accuracy phase results). These changes were:
\begin{itemize}
    \item Correction of electron scattering: due to an incorrect unit setting in the multiple scattering module, the strength of multiple Coulomb scattering had been overestimated in the Accuracy phase dataset, this affected at maximum of 0.5 $\%$ of all particles in the first phase.
    \item Correction of the virtual thickness of the strip modules: in the cluster size calculation of strip clusters a wrong silicon thickness was used initially.
    \item Correction of the beam spot size: the longitudinal beam spot size, i.e. the luminous regions where proton-proton interactions could occur, was corrected from 5.5 mm to 55 mm, which corresponds more accurately to the expected parameters of HL-LHC conditions. 
\end{itemize}

The format of the dataset remained unchanged: the output was organised in a set of comma-separated text files that were organized per event containing both simulated data and ground truth. The ground truth was only available to the contestants for the training dataset.

\subsection {Throughput score}
\label{sec:score}
The goal of this competition is to combine high accuracy and high speed, therefore a score combining both was required.
The experience of the first Accuracy competition has shown that the accuracy score defined as the weighted fraction of points correctly assigned (see \cite{TrackMLAccuracy2019} Eq.~1) was very effective, as the best algorithms in terms of the accuracy score were still the best in terms of the more detailed analysis. A random algorithm has a score 0 and a perfect algorithm has a score of 1; top algorithms in the Accuracy phase reached score just above 0.9 . One slight modification was introduced, which was to remove from the computation of the score trajectories stemming from secondary particles, i.e. particles that do not originate from the primary beam-beam interaction, but from either subsequent particle decay or interaction of primary particles with the detector material. These secondary tracks are not originating close to the origin and are largely less interesting from the point of view of physics. Not considering them eases somewhat the task of the algorithms and reduce their complexity.
 
 The speed is defined to be $t$ the average time per event (in second) as measured on the test dataset, on the allocated resources in the docker environment as detailed in Sec~\ref{sec:codalab}.
 
 The accuracy, $S$, and speed, $t$, were measured on a test dataset of 50 events (instead of 100 for the Accuracy phase), to limit the resource usage on the platform, and also because from preliminary tests the variance of the two quantities appeared to be limited.
 
 In addition, the resource usage in terms of the maximum allowed time was set to $t_{max}=600$~s per event. In order to exclude possible extremely fast algorithms with mediocre accuracy (which would be useless from the point of view of physics), a minimum accuracy of $S_{min}=0.5$ is required.
 
The overall score is given by the formula:
\begin{equation}\label{eq:score}
\sqrt { \log \left( 1 + \frac{t_{max}}{t} \right)    \times ( S - S_{min})^2  }
\end{equation}
The score is an unbounded dimension-less positive number, the higher the better. A particular score value can be reached by different combinations of speed and accuracy, these lines of equal score can be seen on~Fig.\ref{fig:ScoreEvolComp}. The goal of the challenge was to encourage participants to reach the best compromise between accuracy and time, leading towards values in the bottom right corner of the chart. The formula was defined based on the finding of the Accuracy phase, and a poll on the inference speed of their algorithms (which then was not a ranking criterion), which ranged between 100~s per event and a full day. The target expectations for solutions of the Throughput challenge was to yield algorithms with a throughput execution time of 10s per event. Nevertheless, as it will be shown, the formula behaved well also with sub-second algorithms that were submitted.

\subsection {Codalab implementation}
\label{sec:codalab}

The first Accuracy phase of the TrackML competition was hosted by Kaggle, the well-known competition platform. For the Throughput phase, the algorithm had to be run within a fully controlled environment, so that the execution speed could be measured. This requirement was not possible on the Kaggle platform at the time.
The Throughput phase was implemented on Codalab, a platform popular for scientific competitions, which allows for a more customised setup.

Participants had to develop their algorithms on the training dataset (including the ground truth) provided. As for the Accuracy phase, a library\cite{TrackMLlibrary} was available to them in order to evaluate their score. Then they had to prepare and upload the inference part of their algorithm to the platform. 
The platform then runs the algorithm on the public test dataset, statistically identical to the training dataset, but without the ground truth. The execution time and accuracy score, together with the overall score obtained with Eq.~\ref{eq:score}, are measured and reported on the public leaderboard (see Table~\ref{tab:leaderboard}) .
It should be noted that in this Throughput phase, contrary to the Accuracy phase, the public test dataset is undisclosed to the participants, it is ``public'' only because it is used for the public leaderboard. As commonly done in such competitions, in order to avoid that participants overtrain on the public leaderboard score, the overall score was reevaluated after the end of the competition on a private test dataset, statistically identical to the public one.

The overarching goal of the competition was to foster algorithms and ideas from a broader community than the high energy physics community.
In the years the competition was designed then run, the workhorse for large scale processing of LHC experimental data has been single cores CPU with 2GB~RAM.
In practice, the single cores are made available as virtual machines instantiated on a variety of physical hardware in world-wide distributed computing centres. Meanwhile, the community has been moving towards multi-core computing and GPUs are increasingly used in specific contexts.  As a compromise between current and future HEP computing landscape, the resources used to evaluate an algorithm were 2 CPU cores with 4GB of RAM. A participant using a traditional single thread algorithm would lose a factor 2 in speed which would be a small handicap given the logarithmic dependency in speed in Eq.~\ref{eq:score}. We deliberately did not choose to favour a higher level of parallelism in order not to skew the competition towards a parallel-computing coding competition. 

Two physical machines have been dedicated to the challenge, each having two Intel Xeon Processor E5-2650 v4 @2.20 GHz with hyper-threading, each processor having 12 physical cores. This is a typical processor used on the LHC Computing Grid; it has been benchmarked to 10.26 HS06 unit~\cite{HS06gridpp}, a benchmark routinely used for High Energy Physics computing.

Each participant had to implement their algorithm within a provided skeleton.
The skeleton is a small Python class \emph{Model} for which C/C++, 
Python and R bindings examples were given in the starting-kit.
The code is then run by the platform within a 
docker\cite{merkel2014docker} environment limited to 2~CPU and 4~GB of 
memory.
The C/C++ code could be compiled at home and shipped with the binaries 
since the docker was publicly available (which was the method used for almost all submissions).
Before being evaluated, the participant's \emph{Model} could be initialised by compiling source code,  loading data, etc.

The skeleton loops over the input data one event at a time, call the participant's code and writes out the solution found. The time measured is the wall clock time spent in the participant's code (so that all the overhead, in particular in I/O, is not included). To avoid extremely slow submissions using up resources, a pre-test was first done on one single event with a time limit at 600~s. If this pre-test was successful, the measurement was made on the 50~events public test dataset. 

Thorough tests prior to the competition determined that the time evaluation was reproducible to within 2\%, which was independent on possible other evaluations running  concurrently on the same physical machine. This small variance could have, in principle, impacted the ranking of the participants.
Hence, it was decided, and participants were warned about it, that the final measurement of the time of a submission would be done after the end of the competition with a repetition of 10 runs on the final private test dataset. When this was done, no unexpected discrepancies were seen, even though the diversity of code tested was larger than in the tests run prior to the competition.

A standard docker environment was provided including typical libraries. Participants could, in addition,  install on-the-fly libraries from internet, which access remained open to the worker node. However, in practice,  participants preferred to directly ship additional libraries with their own software.
Execution logs were not made available to the participants, as they could have been used to obtain insights on the undisclosed test dataset. Otherwise, there was no thorough attempt to eliminate all possible methods of hacking. Hacking was obviously forbidden in the competition rules participants had to agree to. In addition, the prizes were conditioned to a full release of the source code which was scrutinised (for the top participants) at the end of the competition.

\section {The competition as it happened}
\label{sec:competition}

The TrackML Throughput competition opened on 3~September~2018 a few weeks after the end of the Accuracy competition on Kaggle on 10~August~2018. It was initially due to  on  18~October~2018, but given the initial lack of competitors, it was extended until 15~March~2019.

\begin{table}[htb]
    \centering
    \begin{tabular}{|c|c|c|c|} \hline
    participant     & score & accuracy (\%) & speed (s/event)   \\ \hline
    sgorbuno        & 1.17  & 94.4 & 0.56  \\ 
    fastrack        & 1.11  & 94.4 & 1.11 \\ 
    cloudkitchen    & 0.90  & 92.8 & 7.28 \\
    cubus           & 0.77  & 89.5 & 13.5 \\
    Taka            & 0.59  & 87.5 & 53.4 \\
    Vicennial       & 0.56  & 81.5 & 25.4 \\
    Sharad          & 0.29  & 67.4 & 38.0 \\
    \hline     
    \end{tabular}
    \caption{TrackML Throughput competition leaderboard}
    \label{tab:leaderboard}
\end{table}

\begin{figure}[ht]
\centering
\includegraphics[width=0.48\textwidth]{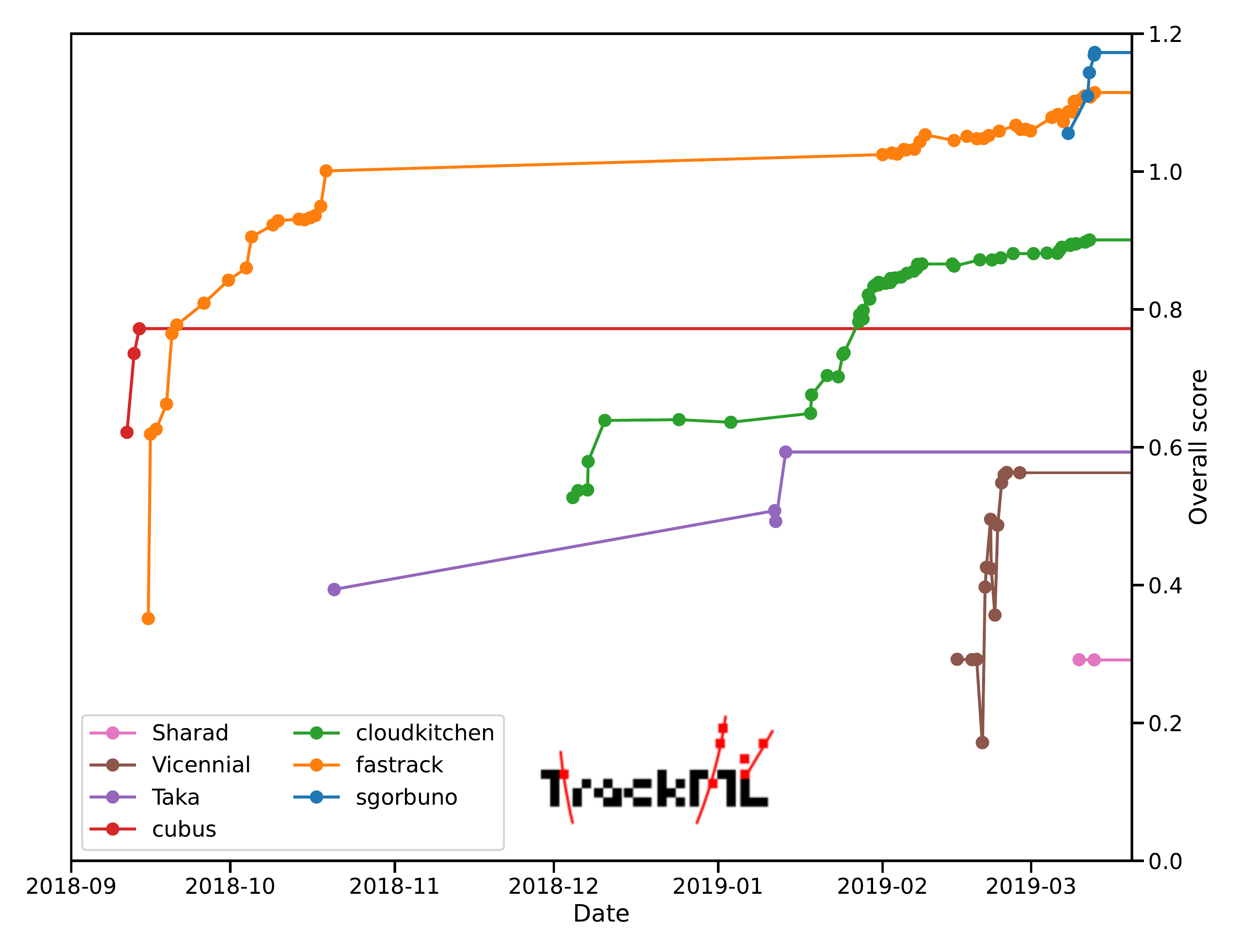}
\caption{
TrackML Throughput phase participants overall score evolution as a function of the date of submission. \texttt{sgorbuno} is Sergey Gorbunov (see Sec.~\ref{sec:gorbunov}), \texttt{fastrack} is Dmitry Emeliyanov (see Sec.~\ref{sec:emeliyanov}) and \texttt{cloudkitchen} is Marcel Kunze (see Sec.~\ref{sec:kunze}).
}
\label{fig:MixedScoreEvol}
\end{figure}

\begin{figure}[ht]
\centering
\includegraphics[width=0.48\textwidth]{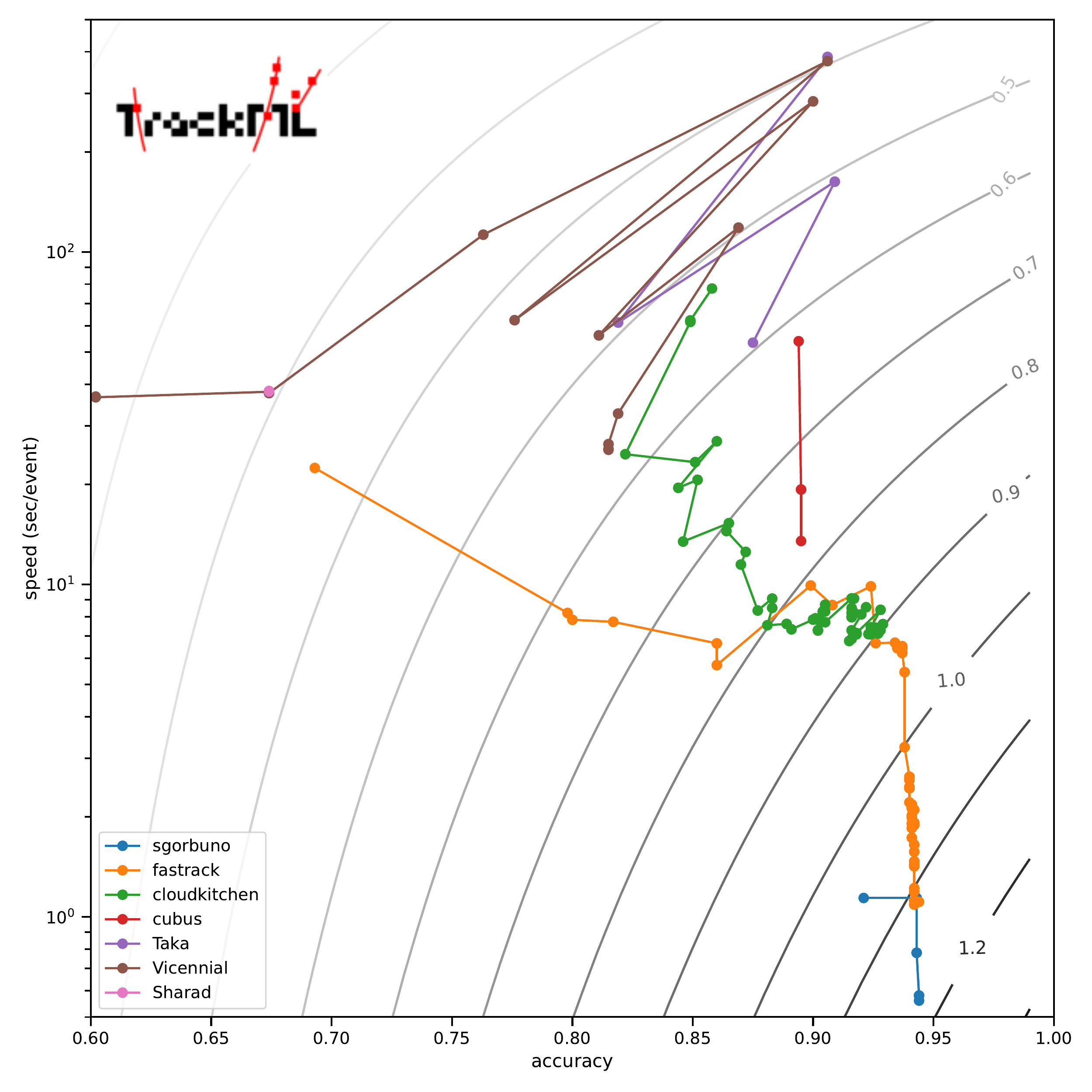}
\caption{
TrackML Throughput phase participants score evolution. 
The horizontal axis is the mean accuracy over the 50 test events, and the vertical axis is the average computation speed per event.
The total score, function of both variables, is displayed in grey contours.
Each colour/marker type corresponds to a contributor, the lines help to follow the score evolution. \texttt{sharad} only made a single contribution, identical to the first point from \texttt{Vicennial}.
}
\label{fig:ScoreEvolComp}
\end{figure}

The leaderboard is shown in Table~\ref{tab:leaderboard}. As detailed in Sec.\ref{sec:score}, participants obtained a non zero score only if their submission could achieve more than 50\% accuracy in less than 600~s per event.
In the end, only seven contributors achieved non zero scores;
their score evolution as a function of date of submission is summarized on Fig.~\ref{fig:MixedScoreEvol} and on Fig.~\ref{fig:ScoreEvolComp} as a function of the accuracy and computation speed.
From the shade of the blue curve, it can be seen that the competition winner, \texttt{sgorbuno}, made a late entry in the competition, with only four submissions that earned him the title.

The number of participants to the Throughput phase has been rather low, especially considering that there were 648~teams participants to the Accuracy phase on Kaggle, which could in principle have carried on to the Throughput phase. In hindsight, this has been understood to come from a combination of factors:
\begin{itemize}
    \item the lower popularity of Codalab compared to Kaggle, where people can earn points across competitions.
    \item the complexity of the problem.
    \item the perceived necessity to write optimised C++ code when a typical Kaggle participant is more familiar with python.
    \item the threshold of less than 600~s per event and more than 50\% efficiency, it was already non-trivial to have a non zero score
    \item despite all the efforts to document and streamline the procedure to submit a solution, it still required a larger commitment than for a typical Kaggle competition.
    \item the fact that we did not provide the log files made debugging rather difficult to the participants.
\end{itemize}

Nevertheless, the fewer number of participants was more than compensated by the high quality of the top three participants (see Fig.\ref{fig:ScoreEvolComp}), who all obtained better than 90~\% accuracy with an execution time up to 0.5~s, compared to an initial goal of better than 10~s per event.
After the end of the competition, all participants made their documented software available, which was scrutinized. The score was re-evaluated on the private test dataset, which confirmed the score from the online leaderboard. Hence the final rankings confirmed the online one.

 The original idea was that the algorithms developed in the Accuracy phase would be optimised and adapted to the second phase, not necessarily by the same participants. This was not enforced in any way but, and it is largely what has happened:
\begin{itemize}
\item Sergey Gorbunov (pseudonym \texttt{sgorbunov})  rank~1 in the Throughput phase had obtained rank~3 in the Accuracy phase (with pseudonym \texttt{Sergey Gorbunov}) 
\item Dmitry Emeliyanov (pseudonym \texttt{fastrack}) rank~2
in the Throughput phase had obtained rank~4 in the Accuracy phase (with pseudonym \texttt{demelian})
\item Marcel Kunze (pseudonym \texttt{cloudkitchen} rank~3 in the Throughput phase ) used as a starting point the algorithm of \texttt{TopQuark}, rank~1 in the Accuracy phase, and has largely augmented it
\item the algorithm of \texttt{outrunner}, rank~2 in the Accuracy phase, was quite innovative but very slow, a full day per event, so was not seen promising enough to be recycled in the Throughput phase. 
\end{itemize}

\section {Algorithmic performances}
\label{sec:performance}

\begin{figure}[ht]
\centering
\includegraphics[width=0.48\textwidth]{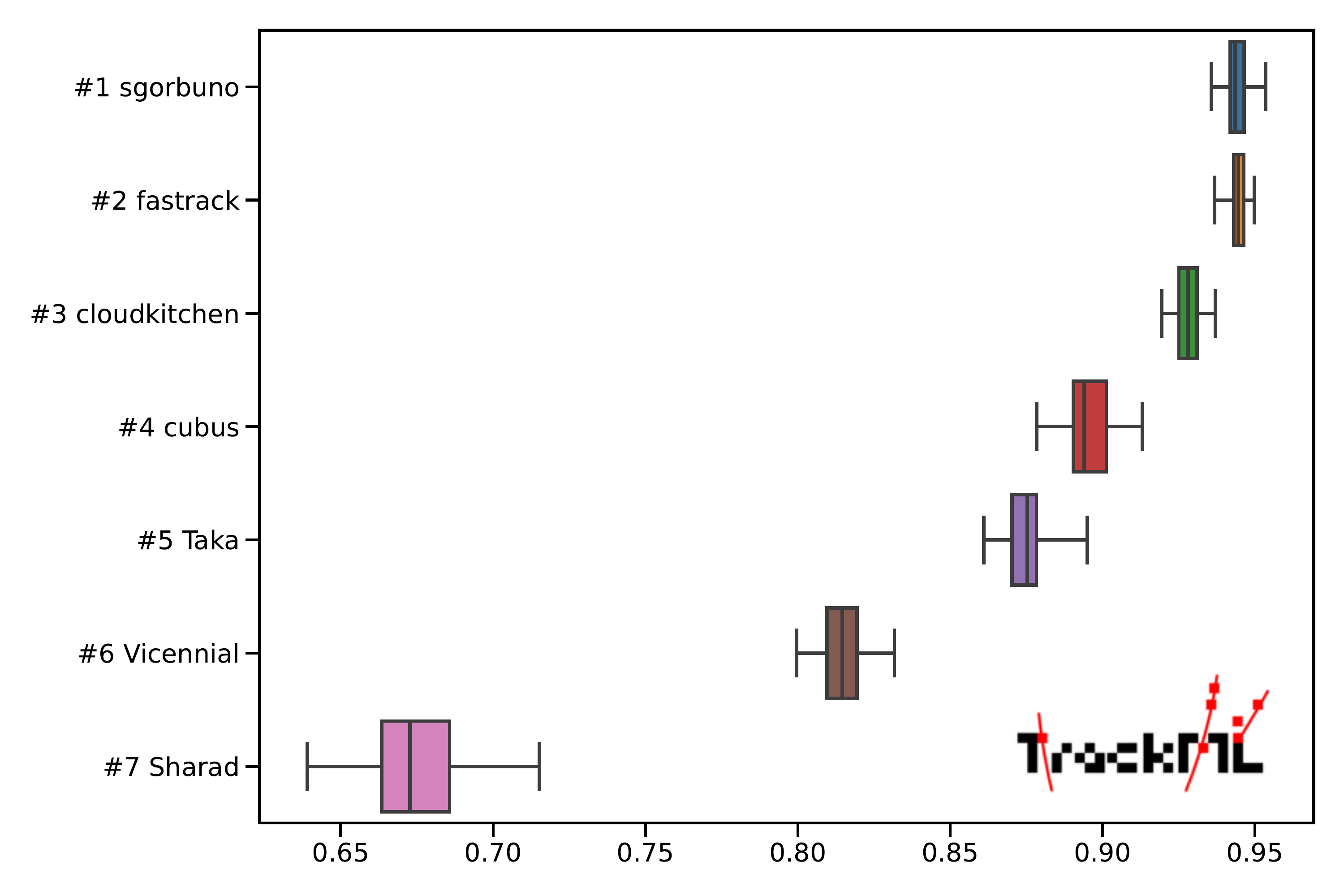}
\caption{
Box plot of the per-event accuracy score on test events for the top participants. The whiskers indicate the total range, the box the quartiles of the individual distributions.
}
\label{fig:LBBoxPlot}
\end{figure}

In this section, a thorough investigation of the performance of the highest ranking algorithms is discussed, as was done for the Accuracy phase in ~\cite{TrackMLAccuracy2019}.
The box plot on Fig.~\ref{fig:LBBoxPlot} indicates the accuracy score on the 50 event test dataset. Interestingly, the accuracy follows the general ranking, indicating that little compromise was made in optimising the algorithms.
Only the first two candidates have very similar accuracy and differ in regards to the speed, as could be seen in Table~\ref{tab:leaderboard}.

Performance assessments of HEP detectors are typically several hundred pages in length, with many histograms assessing the performances from various angles. For practicality, algorithms evaluated in the TrackML challenge are ranked based on a single score, concerning accuracy and speed. In the TrackML Accuracy paper\cite{TrackMLAccuracy2019}, it was demonstrated that the Accuracy score was indeed selecting algorithms which were indeed the best after a more thorough analysis.
This analysis is repeated here for the TrackML Throughput competition, to ensure that the assertion still holds despite the strong speed incentive.
Instead of using the Accuracy score, which is a hit-based efficiency (weighted fraction of points correctly assigned), we use the particle-based efficiency, which is the fraction of particles correctly reconstructed; this quantity is more commonly used in particle physics. A particle is considered to be correctly reconstructed if there is a track sharing more than 50\% of the points with the original particle, as indicated by the ground truth. Contrary to the Accuracy score, this efficiency is not weighted to decrease the relative weight of the lower transverse momentum (larger curvature) particles. This is the main reason why the particle efficiency is a few per cent less than the Accuracy score. The fake rate (the fraction of tracks that cannot be uniquely attributed to a truth particle, another quantity commonly used in particle physics) has not been studied in depth,  because, given the requirement that one point can only be assigned to one track, the fake rate was found to be very much anti-correlated to the efficiency.

\begin{figure*}[t]
\centering
\includegraphics[width=0.97\textwidth]{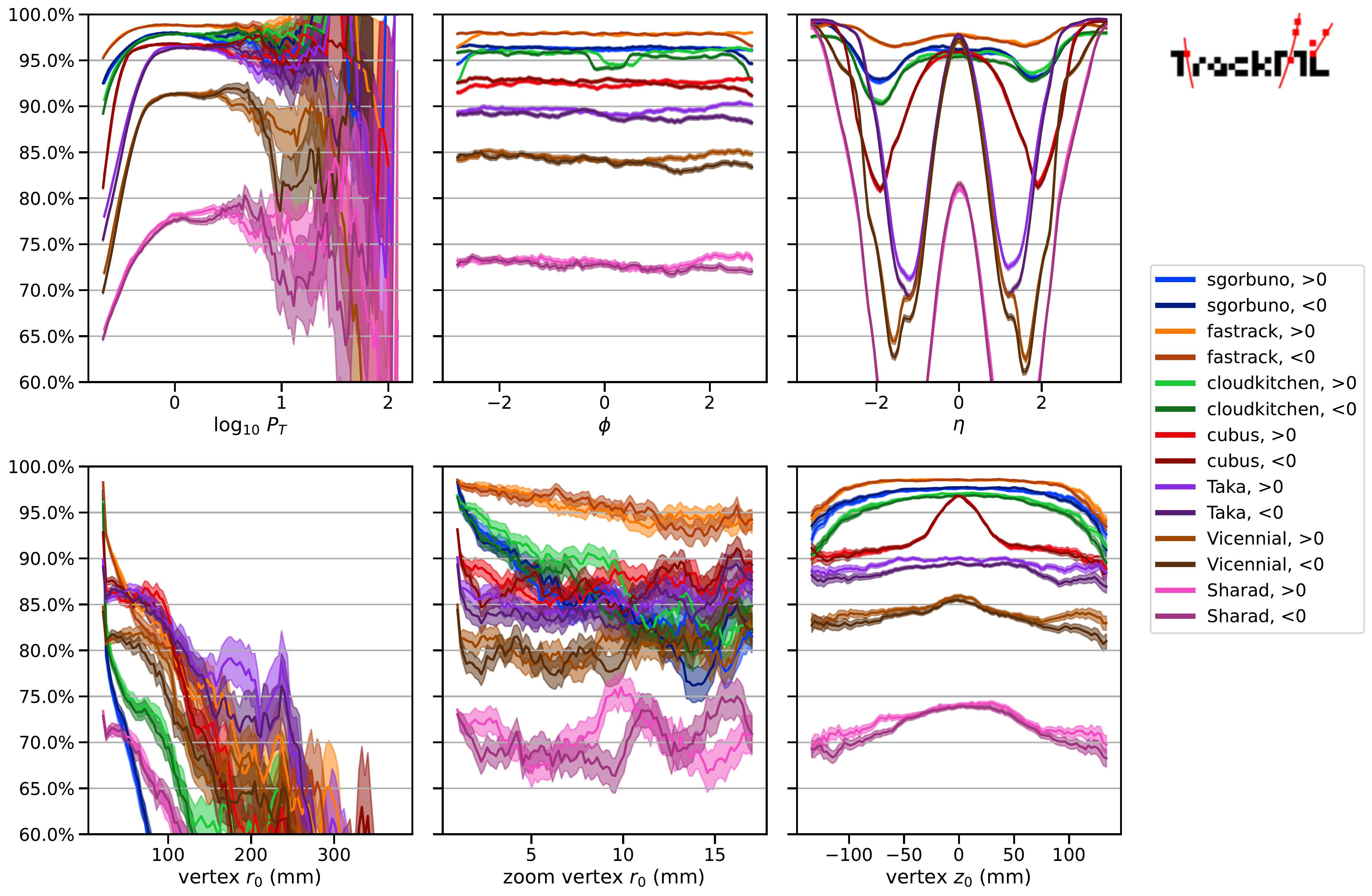}
\caption{
Efficiency as a function of six physical variables ($\log_{10} P_T$, $\phi$, $\eta$, vertex distance $r_0$ from beam axis in mm, zoom on the latter, and vertex beam axis coordinate $z_0$, see text for details) for different participants, each represented by a different colour. Only primary particles are included.
Light-shaded curves are for positively charged particles, dark-shaded ones for negatively charged particles. The band shows the statistical uncertainty on the efficiency measurement.
}
\label{fig:6fig}
\end{figure*}

Fig.~\ref{fig:6fig} displays the efficiency for the 7 best participants as a function of several variables relevant for physical analyses (these variables are obtained from the truth particle):
\begin{itemize}
    \item $z_0$ is the $z$ coordinate of the vertex.
    \item $r_0$ is the transverse distance to the beam axis, $z$, of the particle vertex (creation point)  
    \item $p_T$ (GeV), the transverse momentum, is the projection on the plane perpendicular to the beam axis of the momentum $P$, product of the particle speed by its relativistic mass; for particles of unit charge, it is proportional to the radial component of the particle trajectory.
    \item $\phi$ (rad) is the azimuthal angle (around the beam axis)
    \item $\eta$, the pseudorapidity, is a dimensionless quantity describing the angle of the particle with respect to the beam axis and varying between -4 and 4 for this detector; with $\theta$ the angle in radians, we have $\eta \equiv  \ln ( \tan \theta/2) $ 
\end{itemize}{}

The efficiency curves for the seven participants are well separated. A striking feature is that \texttt{fastrack}'s efficiency is consistently better than \texttt{sgorbuno}'s, despite the two having very close Accuracy score. This is because \texttt{fastrack}'s tracks are typically missing more points than \texttt{sgorbuno}'s, thus lowering its Accuracy score, which is a point-based efficiency.

All algorithms have similar $p_T$ dependencies with a dip at low $p_T$ which correspond to particles with large curvature. Reconstructing these large curvature particles might require to enlarge the search window, at the risk of increasing the number of combinations and decreasing speed; it is also the case that these particles are more difficult to reconstruct because they suffer more material interactions.
The best algorithms are able to mitigate this effect. After a plateau, the efficiency decreases slightly for $p_T$ above 8~GeV. This common feature has not been understood as these particles are almost straight and in principle easy to find. Although this concerns less than a per mil of all particles as can be seen Fig~5 in \cite{TrackMLAccuracy2019}, they can be of high interest from the point of view of physics. This feature was already seen, although less pronounced,  in the Accuracy phase (Fig~13 in \cite{TrackMLAccuracy2019}). It is most likely a side effect of the speed optimisation, which was not noticed by the participants given the very small weight of this region of the phase space in the calculation of the accuracy.

Given the cylindrical symmetry of the detector (see Fig~4 in~\cite{TrackMLAccuracy2019}), the efficiency is expected to be flat according to $\phi$. In general, this is observed.
For \texttt{fastrack} and \texttt{sgorbuno} the efficiency for positively charged particles shows a dip just above $\phi=-\pi$, and another dip for negatively charged particles just below $\phi=\pi$. 
Due to the approximately solenoidal magnetic field pointing along the $z$ axis, positively charged particles turn clockwise, so positive particles starting with $\phi$ just above $-\pi$ are then crossing the $\phi=\pi$ boundary into the region with $\phi$ just below $\pi$ (and the opposite for negative particles). 
The $\phi=\pi$ boundary does not correspond to any concrete geometric feature of the detector so the dips are likely due to a feature of the implementation.
The $\phi$ efficiency curve for \texttt{cloudkitchen} shows a dip at $\phi=0$, shifted between positive and negative particles, which is most likely due to a similar feature when handling the $\phi=0$ boundary.

The efficiency curves as a function of $\eta$ show the evolution of the efficiency as a function of the polar angle, for track close to $-z$ direction ($\eta=-4$), perpendicular to $z$ ($\eta=0$) and then track close to $z$ direction ($\eta=4$). All the curves are symmetric, as expected, and showing a more or less deeper dip around $|\eta|=2$. In these regions, as can be seen Fig~13 in \cite{TrackMLAccuracy2019}, tracks cross the first disks. The best algorithms manage to handle this transition much better than the others.  

The efficiency curves as a function of $r0$ are as expected highest at $r0=0$ since most particles are originating very close to the origin because only primary particles are taken into account in the score and in the efficiency. The efficiency drops rapidly as $r0$ increases because assuming particles are coming from the origin is a strong constraint which increases the speed of the algorithms.

The $z0$ of the primary particles follow a centred Gaussian distribution with a width of 55~mm (this was 5.5~mm for the Accuracy phase). Participants have successfully accomodated for this, and obtained a relatively flat efficiency, except for \texttt{cubus}.  

\begin{figure*}[t]
\centering
\includegraphics[width=0.97\textwidth]{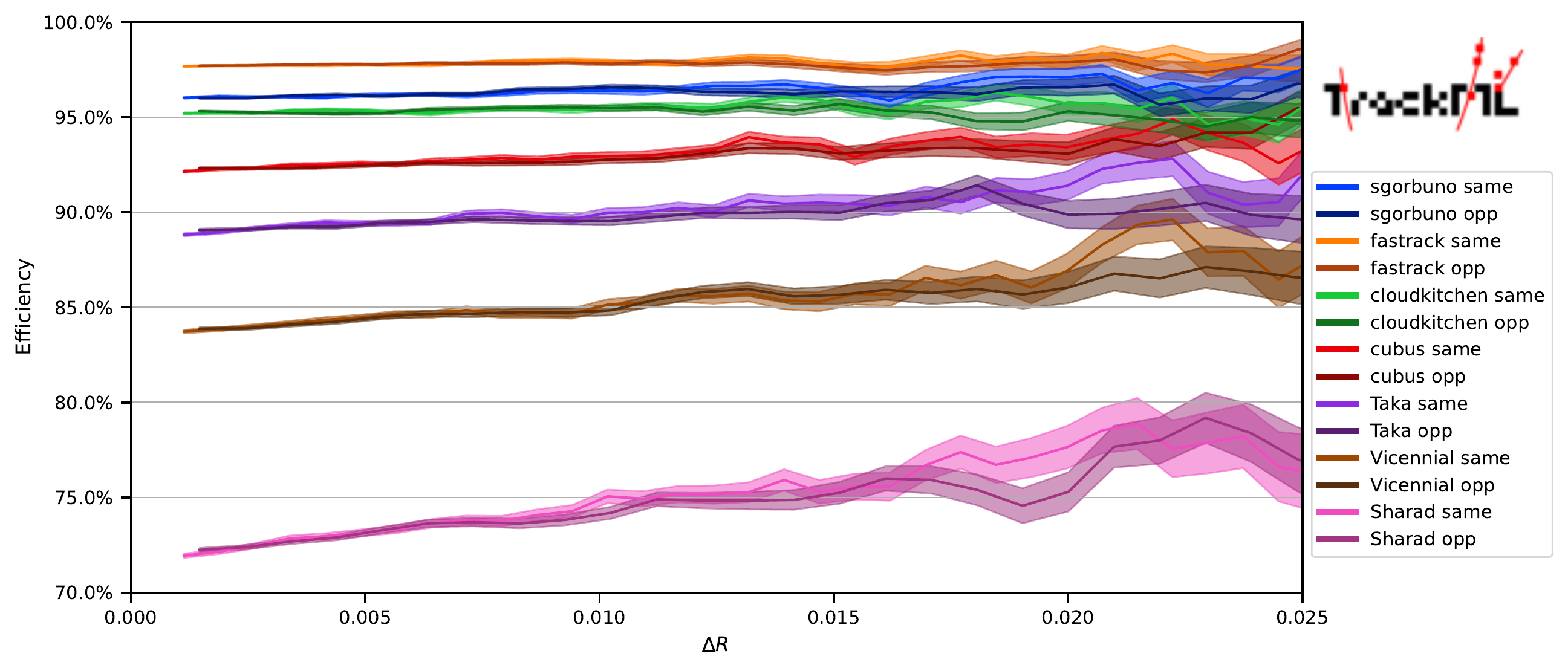}
\caption{
Efficiency as a function of the distance $\Delta R$ to the nearest neighbouring ground truth particle for  different participants, each corresponding to a different colour.
Particles having the same sign as the nearest neighbour are indicated by light-shaded, plain lines, and of the opposite sign with dark-shaded, dotted lines.
}
\label{fig:deltaR}
\end{figure*}

To investigate deeper the quality of the algorithms, the efficiency as a function of the angular separation between tracks was studied. The typical separation variable (commonly used to analyse LHC proton collisions) is defined to be
$$\Delta R =  \sqrt { \Delta \phi ^2 +  \Delta \eta ^2 } $$ (for small values of $\Delta \eta$, $\Delta R$ is similar to the 3D angle in radian).
For each particle, the nearest neighbouring ground truth particle as a function of $\Delta R$ is searched for.
Fig.~\ref{fig:deltaR} shows the efficiency as a function of $\Delta R$ of the nearest neighbour (few particles have a neighbour distant of more than 0.025).
The best three participants achieve a reconstruction efficiency independent of the distance to the nearest neighbour, while the other participant algorithms achieve a slight drop of efficiency for low values of $\Delta R$.
Neighbours of the same charge stay close together for a longer distance as they move away from the origin since they are turning in the same direction. However, there is no visible effect on efficiency whether the nearest neighbour is of the same or of opposite charge, which is a sign of robustness.

\section {Winner : Mikado by Sergey Gorbunov}
\label{sec:gorbunov}
The Mikado approach for the TrackML challenge is a combinatorial algorithm. Its strategy is to reconstruct data in small portions, each time trying not to damage the rest of the data. The idea resembles a Mikado game, where players need to remove carefully wooden sticks one-by-one from a pile without the pile collapsing.

The algorithm performs 60 reconstruction passes with different settings. During the first passes, it only looks for high-momentum  (hence almost straight) tracks within very tight requirements. Found tracks are removed from the detector after each pass, thereby reducing combinatorics for the subsequent passes. The cuts are loosened, and the algorithm runs again. For the last passes, the cuts are very loose, allowing the algorithm to collect all the remaining tracks. 

Despite the high combinatorial factor, the outcome of the first passes is very pure. There are almost no incorrect hit-to-track associations. During the last passes, the algorithm accepts almost everything it finds. Therefore at the latest stage, it collects many wrong hit combinations in addition to the real tracks.

Performing 60 reconstruction passes within a reasonable time is only possible when data access is fast. To do that, the hits from every detector layer are arranged in a two-dimensional grid. The algorithm accesses only those hits which are located within a current search window and the other hits are untouched.

The algorithm uses fixed-size search windows that are different for each detector layer and reconstruction pass. Therefore, tens of thousands of internal parameters need to be tuned. Optimal parameters are not calculated mathematically but trained on the training dataset. The optimising routine is, unfortunately, not fully autonomous and requires manual intervention.

The Mikado tracker shows $94.4\%$ accuracy and takes 0.56 seconds per event.

\subsection{The algorithm}

The algorithm uses a local track reconstruction model. Each time it needs to estimate a particle trajectory, it creates a local helix through three nearby hits that belong to the particle. This three-hit helix is the most flexible trajectory model which follows all local features of a real trajectory. Even though the model uses only a minimal amount of measurements (contrary to classical algorithms which will build a model from all the points already assigned to the track being built), it appears to be accurate enough to predict the particle position on neighbouring layers.

\begin{figure}[bht]
\begin{minipage}[t]{0.4\linewidth}
    \centering
    \includegraphics[width=1\textwidth]{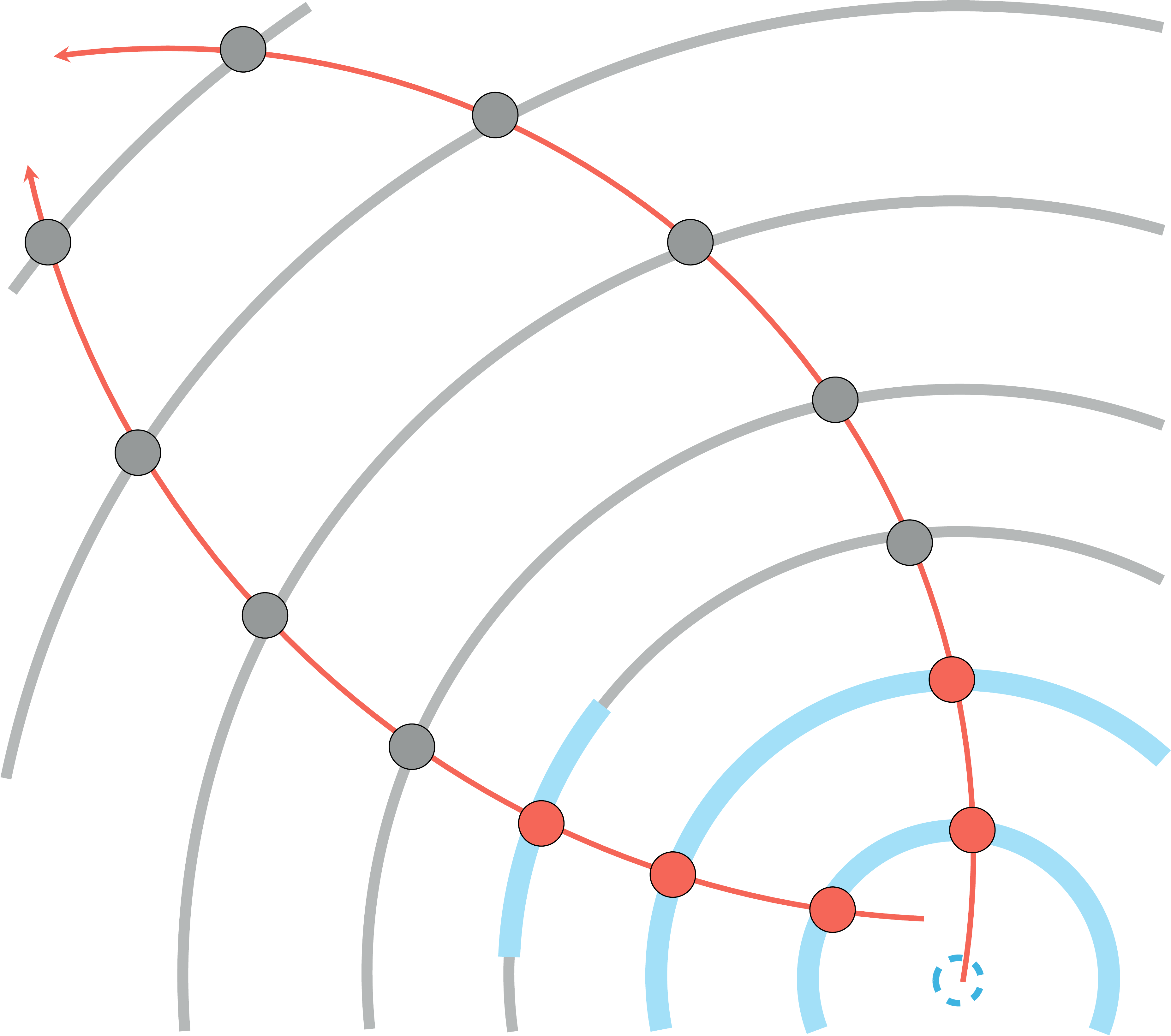}
    \caption{Combinatorial Layers}
    \label{fig:MikadoTrackerFig3}
\end{minipage}
\hfill
\begin{minipage}[t]{0.5\linewidth} 
    \centering
    \includegraphics[width=1\textwidth]{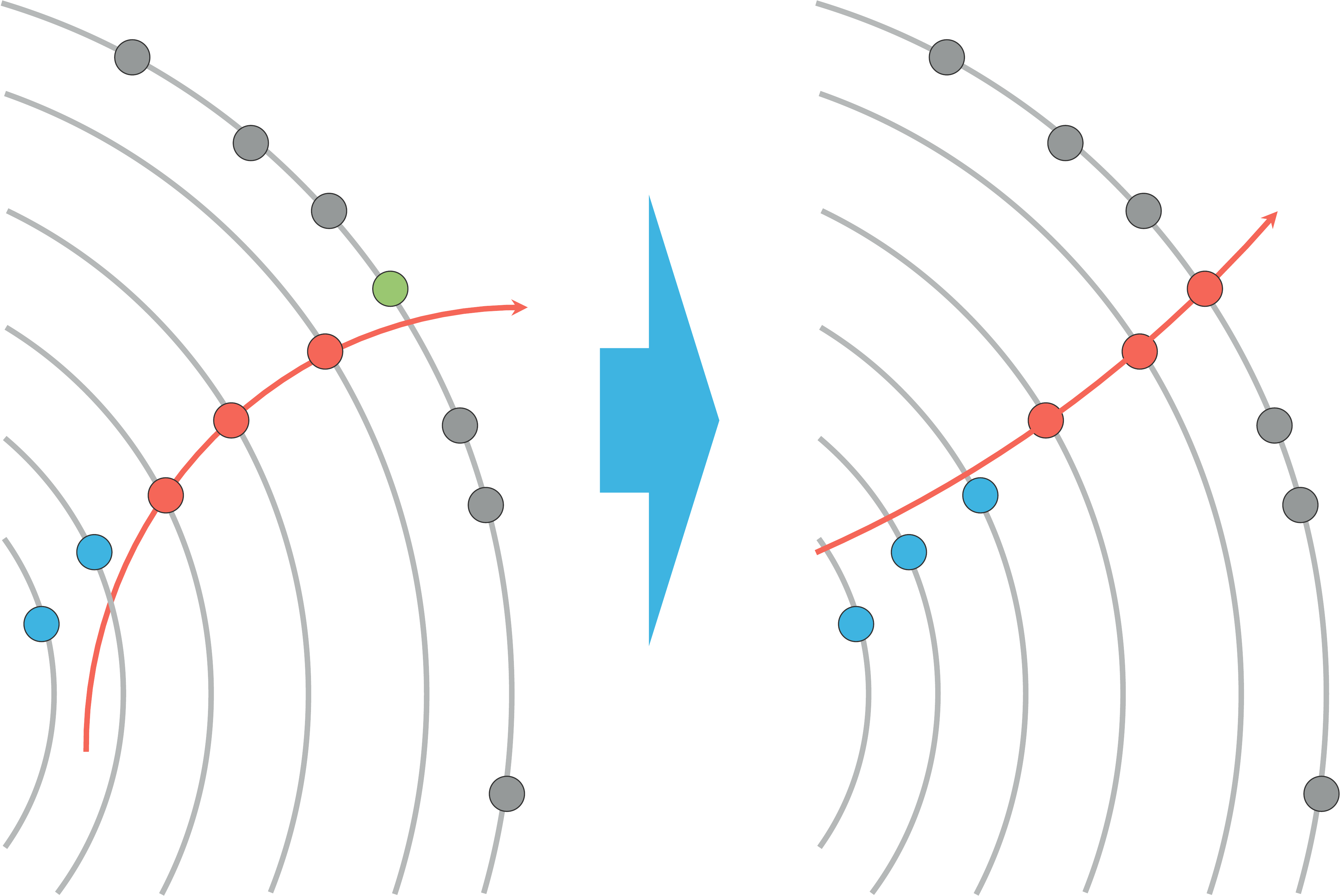}
    \caption{Tracklet prolongation}
    \label{fig:MikadoTrackerFig4}
\end{minipage}        
\end{figure}  

Detector layers consist of many detector elements and they have different orientations in space. For easy navigation to the next modules where to search for more hits, each detector layer is represented as a two-dimensional surface. This surface has two coordinates: a polar angle $\phi$ and the second coordinate $t$, which is equal to $z$ coordinate for cylinders and a radial $r$ coordinate for disks. We project all the hits towards the origin onto their layer surfaces and use their $(\phi,t)$ projections when searching for hits.

For fast data access, there is a regular two-dimensional grid created at each layer that stores the hits in corresponding grid cells according to their $(\phi,t)$ coordinates.

The algorithm consists of three parts. The first part is {\it tracklet construction}. It creates short tracklets on pre-selected detector layers from the hits. The second part is {\it tracklet prolongation}. It creates track candidates by extending the tracklets to other layers and collecting their hits there. The third part is a final {\it selection} of good tracks among the track candidates. To make full use of the two available threads, the code is multithreaded; each thread processes all combinations with one hit on the first combinatorial layer.

The algorithm flow is as follows.

\begin{enumerate}
\item	{\it Tracklet construction}~(see Fig.~\ref{fig:MikadoTrackerFig3})

It is performed on three (optionally, two) selected layers. 
\begin{enumerate}
\item  Every $hit_1$ from the $layer_1$ is considered. Optionally, the first hit can be the origin $(0,0,0)$. 
\item  A straight line which connects the origin and the $hit_1$ is projected to the $layer_2$.
     Within a $(\phi,t)$-search window every $hit_2$ is considered.
\item A straight line, which connects $hit_1$ and $hit_2$ is projected to the next layer, $layer_3$.
     Again, within a $(\phi,t)$-search window every $hit_3$ is considered.
\item A helix of axis collinear to the $z$ axis is constructed on $hit_1, hit_2, hit_3$. In the xy plane, the helix crosses all the three hits, in $Z$ it goes through $hit_2$ and $hit_3$, 
as shown in Fig.~\ref{fig:MikadoTrackerFig8}. 
Then a distance in $z$ of the helix from the $hit_1$ is examined. When it is too large, the hit combination is rejected. Otherwise, the set of three hits is accepted as a tracklet, and the prolongation step starts.
\end{enumerate}

\item	{\it Tracklet prolongation}~(see Fig.~\ref{fig:MikadoTrackerFig4})

The tracklet is prolonged to the next detector layer along its trajectory and the closest hit on that layer is identified.

\begin{enumerate}
\item If the hit is close enough, it is added to the tracklet and the trajectory is recreated using the new hit and two hits from the previous layers. Given that modules can overlap in the same layer, we perform a search for additional hits on the layer within a tiny window around the recreated trajectory. 

\item When there is no good hit found on the layer, when the prolonged trajectory crosses the layer's inner part far from edges, a hit on the layer is considered to be missing.
When hits are missing on more than one layer, the prolongation stops.

\item In certain cases a good closest hit is found, all the additional hits on the recreated trajectory are picked up, but some hits remain in the search area. In this case, the algorithm creates another search branch with a different hit on this layer. The branching is realised in an efficient way with almost no computational overhead.
\end{enumerate}

Once the layer have been processed, the tracklet is extended further until hits on all the layers are collected. Then the tracklet is stored in a list of track candidates and the next tracklet is processed.

\item	{\it Selection of good tracks}

The selection of good tracks is performed by identifying the best one in the list of track candidates. The track should have more hits than the others or at least the same number of hits, but a smaller average deviation of its hits from its trajectory. 
The best candidate is stored as a ``track''; its hits are removed from the detector. 
Then the search for the next best candidate is performed, and so on.  
The selection stops when the best candidate no longer have enough hits.

\end{enumerate}

Once all tracks of the current pass have been founf, the algorithm repeats from step~1, performing the next pass of the reconstruction with a new set of base layers and new search parameters.

\subsection{Fast data access: regular 2D-grid on detector layers}

\begin{figure}[bht]
\begin{minipage}[t]{0.45\linewidth}
    \centering
    \includegraphics[width=1\textwidth]{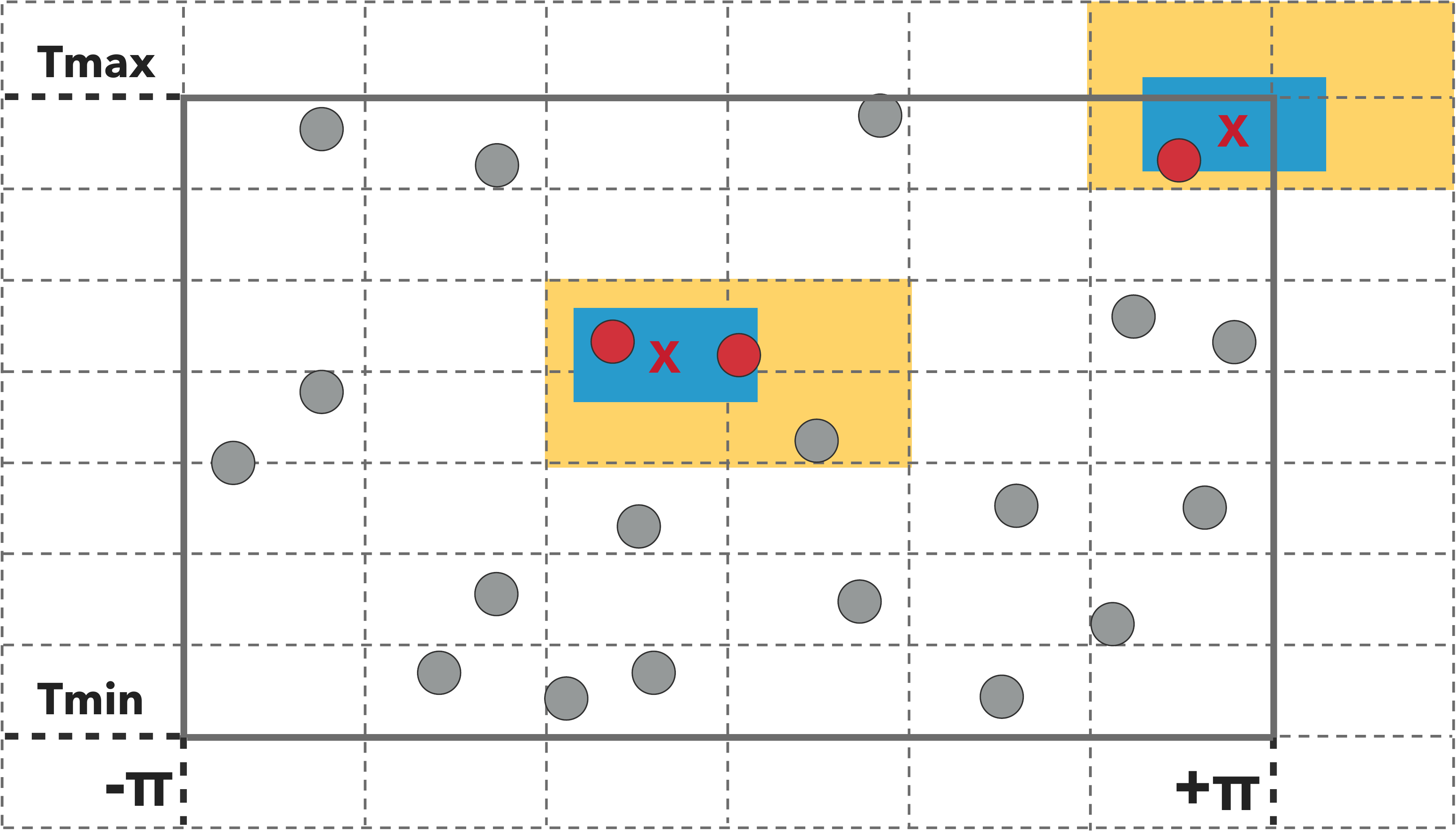}
    \caption{Grid structure for storing hits on a detector layer. 
To find  hits in a blue area, one needs to examine four yellow cells around it. }
    \label{fig:MikadoTrackerFig5}
\end{minipage}
\hfill
\begin{minipage}[t]{0.45\linewidth} 
    \centering
    \includegraphics[width=1\textwidth]{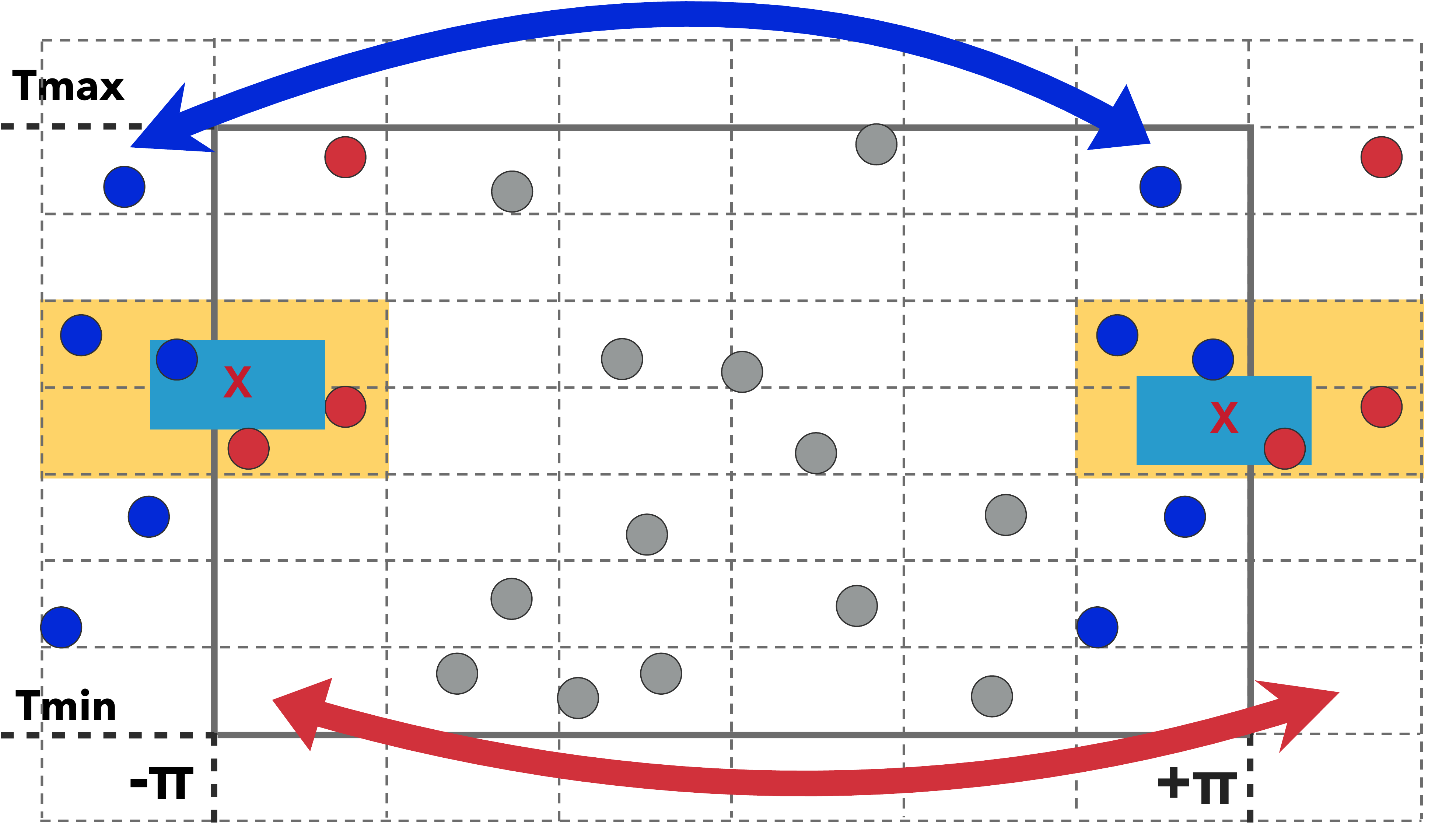}
    \caption{The grid overlap in $\phi$}
    \label{fig:MikadoTrackerFig6}
\end{minipage}        
\end{figure}  

For a combinatorial algorithm, it is crucial to have fast access to data.
For this purpose, hits on every detector layer are stored in cells of a regular two-dimensional grid.

The size of the grid cells is equal to the size of a predefined search window, which is specific for each detector layer in each reconstruction pass.
Search for hits inside the search window is simple. First, one calculates the cell index of the centre of the search area by doing a couple of modulo operations. Then one looks over the hits in four neighbouring cells overlapping with the search window, as it is illustrated in Fig.~\ref{fig:MikadoTrackerFig5}.

To avoid any special treatment of border regions, the grid is surrounded by layers of empty cells. Unfortunately, this technique does not work with the~$\phi$ border at~$\pm\pi$, as this border of the detector surface is purely virtual. 
To handle the~$\pm\pi$ border, we let the grid overlap in~$\phi$. For that purpose, the surrounding empty cells at~$\pm\pi$ are filled with hits from the opposite $\phi$ edge, as it is shown in Fig.~\ref{fig:MikadoTrackerFig6}. The overlap solves the border problem for $\phi$ and covers the~$\pm\pi$ region without introducing unnecessary conditional branches in the code. However a small inefficiency is left as reported in Section~\ref{sec:performance}.

The implementation of the grid is presented in Fig.~\ref{fig:MikadoTrackerFig7}. The grid consists of two arrays: the array of hits $\bf A_1$ and the array of grid cells $\bf A_2$.
Each cell contains only two values: index of its first hit in the array $\bf A_1$ and the number of hits in the cell.
The creation of the grid is extremely fast, created by looping twice over the input hits and twice over the grid cells and performed as follows:
\begin{itemize}
\item initialize the number of cell hits in the $\bf A_2$ array to $0$
\item loop over the input hits and count number of hits in all the cells in $\bf A_2$
\item loop over the cells in $\bf A_2$ and calculate their pointers to $\bf A_1$ according to the number of hits in cells
\item loop again over the input hits and copy them to their places in the $\bf A_1$ array according to their cell number. This is done with a deep copy in order to avoid multiple reference look-up during the combinatorial search.
\end{itemize}
The efficient access to the data makes the algorithm fast and allows many reconstruction passes to be performed within a reasonable computing time.

\subsection{Physical trajectory model and the magnetic field}

\begin{figure}[bht]
\begin{minipage}[t]{0.45\linewidth}
    \centering
    \includegraphics[width=1\textwidth]{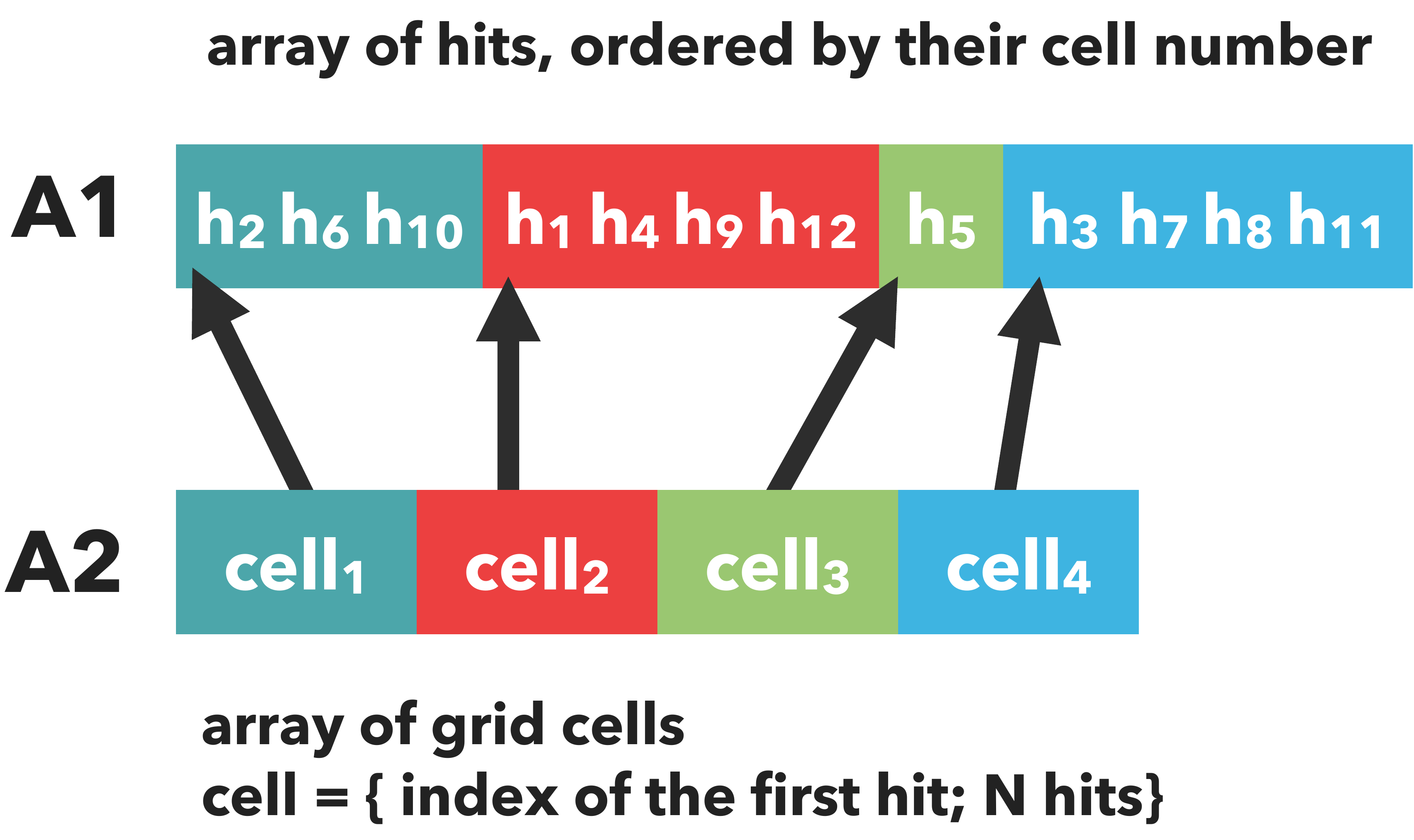}
    \caption{Implementation of the grid}
    \label{fig:MikadoTrackerFig7}
\end{minipage}
\hfill
\begin{minipage}[t]{0.45\linewidth} 
    \centering
    \includegraphics[width=1\textwidth]{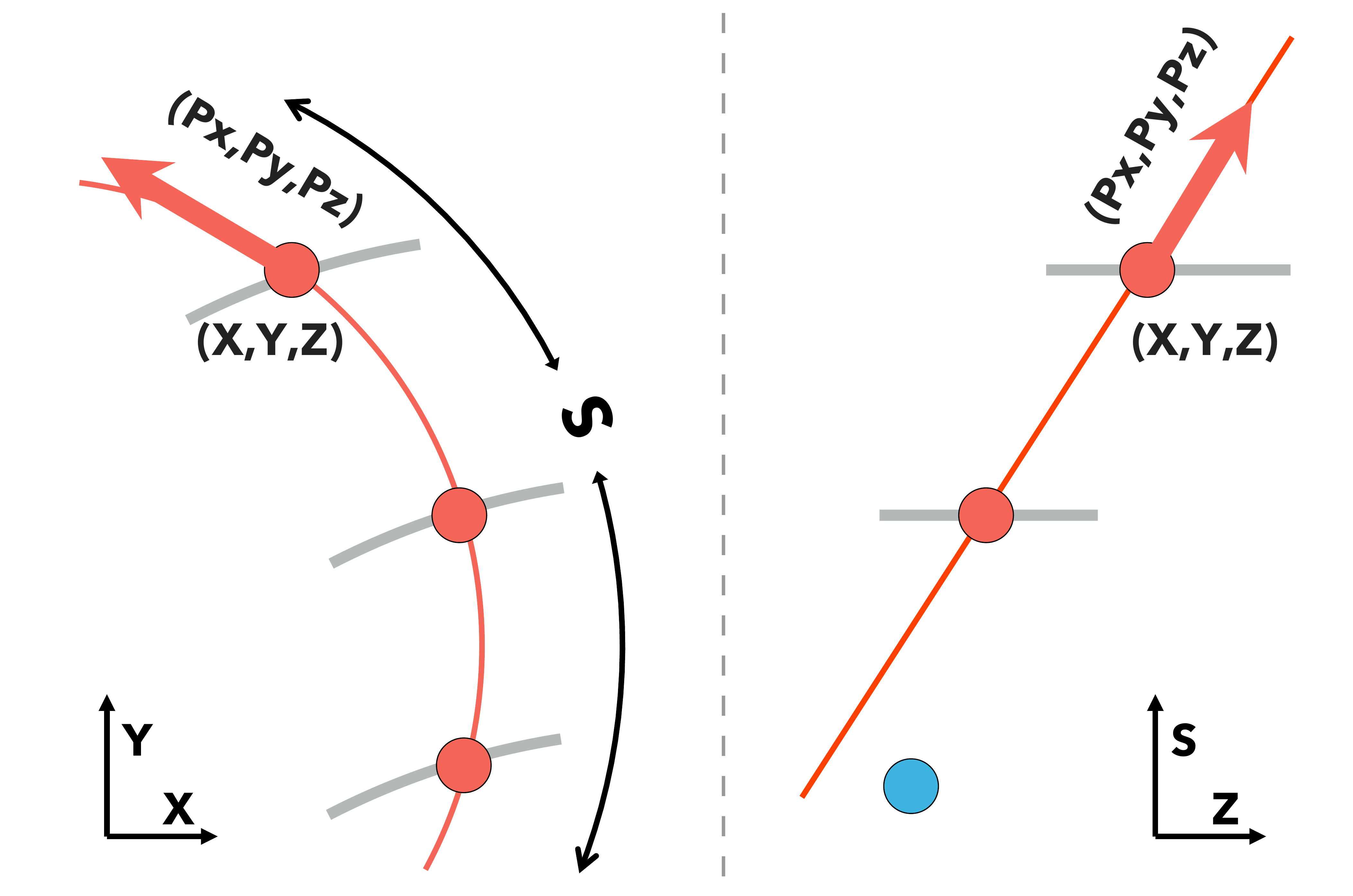}
    \caption{Physical track model}
    \label{fig:MikadoTrackerFig8}
\end{minipage}        
\end{figure}

The magnetic field changes significantly from layer to layer, which means that trajectories  deviate from a mathematical helix.
In order to account for this feature, the physical trajectory model $(x,y,z,p_x,p_y,p_z,q)$ is used. It is presented in Fig.~\ref{fig:MikadoTrackerFig8}. Here $(x,y,z)$ is the spatial position of a trajectory point, $(p_x,p_y,p_z)$ are the three components of the particle momentum, $q =\pm 1$ is the charge.

Even though a description of the magnetic field was not provided in the competition, an average field on each layer can be constructed using the ``truth'' data and approximating this field with a simple polynomial model. The polynomial field is calculated at initialization time for each detector layer and stored in a geometry file. 
At the beginning of the event reconstruction, the approximate field value is calculated for every hit using the above polynomials and stored directly in the hit data structure.
Three different approximate field values are actually used: one for the construction of the local helix, one for the inward prolongation of the helix and one for the outward.

\subsection{Parameter tuning}

To achieve the best result, one has to tune all thousands of algorithm parameters simultaneously, maximizing the overall score Eq.~\ref{eq:score}. But due to a lack of computing resources, a step-by-step optimization is performed instead. Each reconstruction pass is adjusted individually, optimizing the result of the partial reconstruction after that pass. 

The overall score Eq.~\ref{eq:score} is inapplicable in this scheme. Therefore each pass uses its own optimization criterion, which is a compromise between the number of tracks found at that pass and the  purity of these tracks. The pass optimization is performed several times with manually adjusted criteria, set depending on the results. 

Within a reconstruction pass and the chosen optimization criterion, the parameters are adjusted automatically using a primitive gradient following method. The reconstruction time is monitored manually and not explicitly included in the optimization.

\subsection{Outlook}

As the Mikado tracker performs fast hit search within predefined search windows, it has tens of thousands of internal parameters to tune (e.g. size of the search windows). This situation is not typical for traditional track finders, where the search windows are estimated individually for each track using relatively involved trajectory extrapolation with uncertainties. These internal parameters have been tuned semi-automatically on the training dataset. 

\section {Runner up : FASTrack by Dmitry Emeliyanov}
\label{sec:emeliyanov}
The FASTrack (Filter and Automaton for Silicon Tracking) algorithm won the second place in the throughput phase of the TrackML competition with an accuracy of 0.944, a processing time per event 1.11 seconds and overall score 1.1145. After several post-competition improvements, the final accuracy of the algorithm was 0.948 and projected time per event about 0.8 seconds. The memory consumption of the algorithm itself was approximately 0.6~GB and when the algorithm was run in the TrackML docker environment the overall memory consumption was 1.4~GB. 

\subsection{Algorithm summary}

The FASTrack algorithm is based on the following key ideas and techniques:
\begin{itemize}
    \item hit clusters shape (numbers of cells in $u$- and $v$-directions on the module plane) are used to predict the intervals of track inclination angles and save CPU time by avoiding hit combinations with parameters incompatible with the prediction;
    \item the track segment-based track following network is used with an embedded Kalman filter for fast discovery of track candidates;
    \item limited usage of the Kalman filter-based combinatorial track following for missing hits search and track extension to areas not covered by the track following network such as long strip in the outer volumes
\end{itemize}

The track finding is organized as a multi-stage process. There are three stages: the first finds higher momentum central tracks (coming from the interaction region along $z$-axis), the second finds lower momentum central tracks, and the third stage targets the remaining tracks. Once all stages are completed, the output track collections are concatenated and hit labels are generated. In order to create a unique “hit-to-track” assignment all reconstructed tracks are sorted in accordance with their quality and assigned increasing integer track indices (track Ids) so that the best track has the smallest track Id. Then a hit is assigned a track Id only if the hit is not already assigned to another track with a smaller track Id.

\subsection{The algorithm description}

The algorithm starts by arranging input hits into circular “bins” in all the detector layers. The bin widths are calculated in accordance with a uniform $\eta$ binning to guarantee approximately the same number of hits per bin. The width of each $\eta$-bin is 0.2. All hits in the bins are sorted along increasing value of $\varphi$ (azimuthal angle). 
Next, the hits in each layer are clustered into nodes in order to group the hits which likely belong to the same track but are located in different modules on the same layer. The nodes are used for the actual track finding while hits are subsequently used for more precise track fitting.  
After the clustering, the nodes are pre-selected for subsequent track segment creation on the basis on their cells parameters (number of cells along u- and v-directions).  For each selected node an interval on  $\tau = \cot\theta$ (where $\theta$ is the track inclination angle w.r.t. $z$-axis) is obtained using a lookup table which relates the min/max values of $\tau$ to the number of cells in $v$-direction. 
The nodes are connected and track segments are formed in accordance with the layer linking scheme trained on data. For example, the following scheme record for a pair of layers

 8004, 8002, 0.876002

\noindent means that layer 2 of volume 8 is connected to layer 4 of volume 8 and the average amount (called “flow”) of track score carried through this connection is 0.876002. By definition, the initial "flow" emanating from the interaction region is 1.0. The "flow" parameter is used to characterize the importance of layer connections.

To facilitate parallel processing by OpenMP (needed to make full use of the two available cores), the track segments are created and stored in three separate arrays (Segment Banks).  The segment building algorithm operates on node collections from possible pairs of $\eta$-bins (rings in $\phi$). The bin pairing was trained on data to achieve 0.99 efficiency of track segment finding. The output of the training procedure is a set of paired bins indices stored in a look-up table.

The next step of track finding connects track segments, which share the same nodes and creates the track following network. The network is a directed graph in which the vertices are the nodes containing hits and the edges are connections between the nodes, i.e. track segment. For each vertex, there are two collections of edges: incoming and outgoing. The sense of direction is determined towards the $z$-axis of the detector. The algorithm selects all the vertices with non-empty “In” and “Out” collections and for each “In” edge finds possibly connected “Out” edges satisfying cuts on differences in azimuthal angle, pseudorapidity, and the track curvature. The maximum allowed number of connections is set to 6.

Once the network has been built, the segments interact with their neighbours in the “Out” direction. The aim is to calculate the maximum number (called level) of connections which can be traced from the segment and identify the segments which are likely to be the starting points of long tracks. The implementation of this algorithm employs a cellular automaton (CA)~\cite{Emeliyanov:2002}. The CA is parallelized using OpenMP and operates in synchronous mode. First, the proposal for the new level is calculated for all segments (e.g. if a segment with level = 1 has a neighbour with the same level then the proposal for the next CA iteration is 1+1 = 2). Finally, all segments with proposals which differ from their current states are updated.

The network evolution stops once no more segment level updates can be made throughout the whole network. The algorithm then proceeds with the extraction of track candidates from the track following network. The track extraction starts with the segments with level values equal to the maximum level observed during the CA iterations. The algorithm continues with track extraction until the maximum level of remaining segments drops below the stage-dependent threshold (4 for the first stage, 3 and 2 for the second and third, respectively). 

The track extraction is basically a segment-by-segment track following process which is implemented as a recursive “depth-first” graph traversal. In order to quickly reduce the number of traversed combinations, a simplified Kalman track fit is embedded in the recursion. The track fit estimates the track $\cot\theta$ in the $rz$-projection and track $\varphi$ and $d\varphi/dr$ or $d\varphi/dz$ in the $r\Phi$-projection. To speed up the calculations, the track fit does not use any magnetic field description but instead, models the track evolution in $r-z$ as a random walk (caused by the detector material effects) and as the Ornstein-Uhlenbeck (AR(1)) process ~\cite{Ornshtein:1930} in the $r\Phi$ projection which emulates gradual, trend-like, change in the track azimuthal direction under the influence of the magnetic field. 

The more precise track fit of the extracted track candidates is performed using the Kalman filter algorithm which employs the 3rd order Runge-Kutta track parameter a Jacobian extrapolator and a fast approximation for the non-ideal solenoidal magnetic field~\cite{FastField:2015}. The parameters of the solenoid (field in the centre, half-length, and the aspect ratio = radius/half-length) were learned from the data by the tracking efficiency maximization during a hyper-parameter scan. 

As many track candidates share the same hits, some of the tracks are merged and removed in a clone removal procedure. All tracks are sorted in accordance to their fit likelihood (the weighed number of layers with associated hits minus penalty on the $\chi^2$ contributions of hits) and then hits are labelled by the track index starting from the best track. In this way, the shared hits are identified and, depending on the fraction of shared hits and the number of competing tracks, a decision is made whether to merge a track with a better one or to delete it.

The merged tracks are re-fitted and extended towards the interaction region and towards the outer long-strip volumes, as they were not used in the segment creation and network building process. The track extension procedure consists of predicting the track trajectory by extrapolation from the first (last) hit on the track and collecting the hits around the trajectory crossing points on detector layers and track update. Any ambiguity in the “hit-to-track” assignment is resolved via the branching track propagation which also employs the Kalman track fit. The number of simultaneously propagated “best” branches is one for the “inside” track extension and three for the “outside” propagation. The track extension procedure can add up to three new hits per layer to account for situations when more than one hit per layer is produced in the overlapping detector modules. 

Since the track extension can cause additional hit sharing, the clone removal procedure is called again. Next, the extended and possibly merged tracks are refitted and the “hit-to-track” assignments are reviewed. Any missing hits found in the vicinity of estimated track positions on the detector layers are added to a track. The hit addition algorithm applies the constraint that a track can have at most one hit per module. 

Finally, all reconstructed tracks are checked for the number of shared hits. If this number exceeds the stage-dependent threshold (e.g. seven for the first stage) the track is discarded. Otherwise, the track is accepted and all the hits on it are marked as assigned so that they cannot be used in subsequent stages of the track finding.

\subsection{Outlook}

The execution time of the algorithm can be improved by massive parallelisation on General Purpose
Graphics Processing Units (GPGPUs). Several parts of the algorithm are already implemented in a thread-safe manner and accelerated using OpenMP directives. By exploiting the track-level parallelism the track fitting parts of the algorithm can be efficiently executed on a GPU since the fast and compact magnetic field model can be implemented as a GPGPU device code. Currently, the “In“ and “Out“ collections of track segments are created independently. It might make sense to group detector layer pairs in such a way that, firstly, all “In“ collections are formed for a particular layer. Then these collections can be analysed to make predictions for the “Out“ track segments. For example, these predictions can be expressed as an interval of track inclination angle in $rz$-plane compatible with the track segments in the “In“ collection. By using this approach one can avoid creating segments which cannot be connected at their common nodes.

\section {Runner up : Marcel Kunze}
\label{sec:kunze}
The algorithm uses artificial neural networks for pattern recognition based on spatial coordinates together with simple geometrical information such as directional cosines or a helix score calculation. Typical patterns to be investigated are hit pairs and triplets that could seed candidate tracks. The training of the networks was accomplished by presentation of typically 5 million ground truth patterns over 500 epochs. The hit data are sorted into voxels organized in directed acyclic graphs (DAG) to enable fast track propagation. In addition to the spatial hit data, the DAGs hold information about the network model to apply, and a $z$ vertex estimate derived from the ground truth. As they combine the data with the corresponding methods the DAGs form a nice foundation to define tasks that can be run in parallel very efficiently in a multi threaded environment. 
There are two sets of graphs: one set covers detector slices along the $z$-axis, the other covers a grid transverse to the $z$-axis. Each set could be used independently, but a clever combination of the two yields the best overall score: The first set is used to seed the pair finder while the second drives the triplet finder. Prior to the execution of the model the DAGs were trained with tracking ground truth of typically 15-25 sample events, yielding a good balance between graph traversal time and accuracy.
The path and track finding is based on inward and outward triplet prolongation in combination with outlier density estimation, as proposed by J.S. Wind (a.k.a \texttt{TopQuark}) in the Accuracy phase \cite{TrackMLAccuracy2019}. With two threads the execution time is on average about 7 seconds per event at 93\% accuracy in the Codalab docker environment.

\subsection{High-level description}
The tracking model has been designed and implemented as a standard C++11 shared library. It may be run using the \textit{main} C++ driver program, or it may be loaded into the python runtime environment using \textit{ctypes}. The architecture comprises a \textit{Tracker} class for data housekeeping and steering, as well as a \textit{Reconstruction} class to implement the algorithms. The data are organized in the \textit{Graph} class that has been designed as a STL-like header file. The neural networks are handled by the \textit{XMLP} class. The \textit{Trainer} class inherits from \textit{Tracker}: it takes care of neural network training. While the training is based on the \textit{Neural Network Objects} \cite{nno1996} and the \textit{ROOT toolkit} \cite{root1996} there is no dependency of the tracking shared library to external packages. Persistence of graphs and neural networks has been achieved by streaming of the objects to corresponding text files.
The program consists of five parts: setup, pair finder, triplet finder, path finder, and track assignment. The setup stage reads all configuration data and initializes the neural networks and graphs prior to processing the first event. The subsequent parts run as threads in parallel for each event, followed by a final serial track assignment to join the partial results into a common solution. The program implements multi-threading by instantiating corresponding reconstruction objects and managing a set of tasks by using a thread-safe stack. The tasks correspond to graphs that hold the corresponding hit data and a set of neural networks to classify the data. While an event is being processed, each thread pops a task from the stack and executes it. Once the stack is empty and all tasks are finished, the first thread continues and combines the partial results into the final assignment of hits to tracks. The track assignment is written to a result file and handed over to the Python frame that delivers it to the CodaLab platform.

\subsection{Scientific details}
\begin{figure}[ht]
\centering
\includegraphics[width=0.48\textwidth]{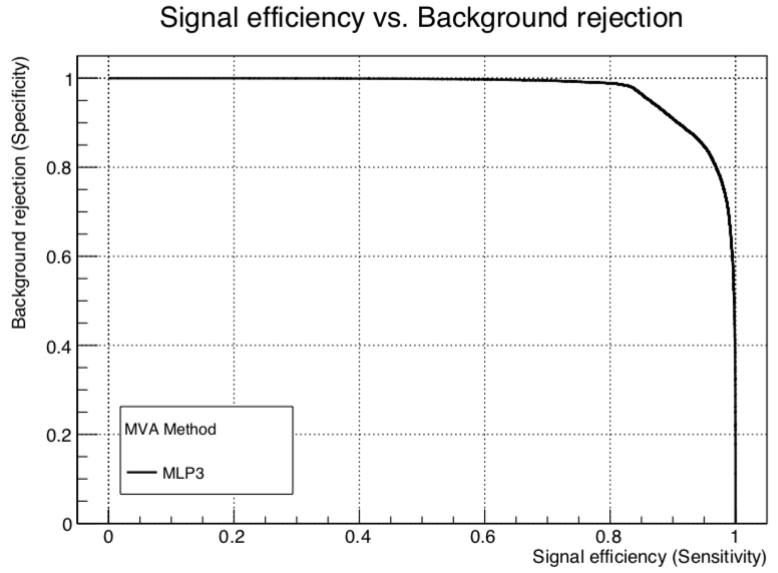}
\caption{
Receiver operation characteristics of triplet finder perceptron \textit{XMLP3}.
}
\label{fig:rocxmlp3fig}
\end{figure}

The model is based on a cylindrical coordinate system ($r_{t},\phi,z$) to describe the hit data. A library of track patterns has been organized utilizing direct acyclic graphs of space elements such that any element has following elements. In principle, the resolution could be chosen on a very fine-granular detector cell level. Although this would yield very accurate results, the resulting graphs tend to grow very large and graph traversal becomes slow. For this reason, a two-dimensional \textit{graphHash} function has been defined to identify a $\phi$/$\theta$ segment for any hit:
\begin{quote}
$i_1 = (int) (\phi_1*0.15*(\pi+\phi));$
\end{quote}
\begin{quote}
$i_2 = (int) (\theta_1*0.1*(5-\theta));$
\end{quote}
where $\theta$ corresponds to asinh$(z/r_{t})$ to flatten the distribution. The constants $\phi_1$ and $\theta_1$ define the granularity of the spatial tessellation. It turned out by tuning that a setting of 12 tiles in $\phi$ and 14 tiles in $\theta$ yielded the best compromise of accuracy vs. speed (i.e. highest overall score). In order to improve execution speed, each tile is bound to a dedicated graph (168 in total). The graphs have been trained by presenting ground truth tracks of typically 15-25 events, which takes about a minute in total for all graphs. 
In addition, a voxel hash function has been defined to identify a hit and its correspondence to a spatial segment:
\begin{quote}
$index = i_1<<32\ |\ i_2<<24\ |\ l<<16\ |\ m;$
\end{quote}
where \textit{$i_1$} and \textit{$i_2$} are the corresponding \textit{graphHash} values and \textit{l} and \textit{m} are the layer and module numbers of the hit, respectively. The use of the shift operator $<<$ in combination with the or $|$ operator allows for very fast construction of the index bit pattern.
There are two sets of graphs: One set covers the two detector slices along the $z$-axis, the other covers the angular grid (tiles). The first set is used to seed the pair finder, the other is used to drive the triplet finder. Each set would work perfectly well by itself, but a clever combination of the two yields the best overall score.
The pair finder utilizes two neural networks, \textit{XMLP1} and \textit{XMLP2} to classify pair candidates. \textit{XMLP1} is an 8-15-5-1 multi-layer perceptron that has been trained with the ground truth cylindrical coordinates of the two hits in addition with the two directional cosines of the hits along the trajectory. \textit{XMLP2} is a 9-15-5-1 multi-layer perceptron that, in addition, takes the helix score as an input as calculated in \cite{TrackMLAccuracy2019}, assuming the origin in addition to the pair. Both networks perform very well in any direction. As it consists of fewer nodes and does not require a vertex calculation, \textit{XMLP1} executes a slightly faster than \textit{XMLP2} at the expense of a few per cent lower accuracy in the central region of the detector system. The final setup therefore combines \textit{XMLP1} for the forward/backward section (“disks”) with \textit{XMLP2} for the central section (“cylinders”). In average there are about 500,000 pair combinations accepted with a cut of 0.15 on the output of  (\textit{XMLP1}) and a cut of 0.55 on the output of (\textit{XMLP2}) yielding an overall tracking score of 99.4\%. The list of pairs is then submitted to the triplet finder. The triplet finder uses a 10-15-5-1 multi-layer perceptron that has been trained with the coordinates of 3 hits plus an additional helix score  (\textit{XMLP3}). On average it accepts approximately 320,000 combinations per event with a tracking score around 97\%. The error rate presenting 100,000 validation patterns reaches about 6-8\% for \textit{XMLP1}/\textit{XMLP2} and around 2\% for \textit{XMLP3}, respectively. Fig.~\ref{fig:rocxmlp3fig} shows the signal efficiency vs. background rejection of \textit{XMLP3} after training 350 epochs of 3.5 million patterns each.

The track finding and assignment is based on inward/outward triplet prolongation in combination with outlier density estimation from \cite{TrackMLAccuracy2019}. It takes care of joining the graph results and yields an accuracy of about 93\%. The track assignment is necessarily executed as a serial task .

\subsection{Interesting findings}
The following interesting findings have emerged during the work for the contest:
\begin{itemize}
\item The training of the neural networks was initially based on pure cylindrical coordinates. It was observed that the input features could be folded in each coordinate due to the detector and event symmetry, thus considerably speeding up the training and reducing the number of patterns. Technically, this octagonal folding is most simply realized by use of the $abs$-function in combination with a $\pi/2$ shift, e.g. $\phi$ is replaced by $abs$($abs$($\phi$)-$\pi/2$).

\item Conventional cuts on the vertex constraint considerably reduce the number of patterns to be processed. Through a simple geometrical estimate of the $xy$ and the $z$ vertices by straight line propagation in the inner layers, the number of patterns to be classified by the neural networks decreases from more than 2,000,000 to about 1,200,000 combinations per event. 
In principle, the vertex determination could be made using a neural network: a preliminary version of the solution integrated a neural network vertex estimate but despite achieving a better accuracy the rather long inference time yielded a reduced overall score.
 
\item The training of the graphs happens once prior to the model evaluation and needs only O(15-25) events to yield optimum results. The graphs are persisted as part of the model. If more events are used during training, more accurate results may be achieved as the track library contains more voxels, however at cost of a longer execution time (and hence a lower overall score).  

\item The accuracy improves by 0.2\% if the graph tasks are organized such that subsequent threads work on neighbouring graphs. This is due to the fact that a track pre-assembly already happens on the thread level prior to the merging of the partial results at the end. In that way, overlapping paths are already being merged on the parallel thread level thus relieving the serial task.
\end{itemize}

\subsection{Outlook}
Great care has been taken to avoid using any low-level detector specific information in the core tracking algorithms in order to keep the algorithm as generic as possible. The neural networks are mainly trained with spatial information. As such the algorithms could be easily transferred to other environments or detectors.

The graph implementation furthermore offers a serialization function that allows a list of tracks to be quickly obtained from the triplets stored in a DAG by recursive graph traversal. This already works surprisingly well in an environment with a lower track density (up to a few hundred tracks).

\section {Conclusion}
\label{sec:conclusion}
The TrackML challenge has been a long running competition series to gather new algorithmic ideas to speed up tracking in the LHC experiments.
After the first round of initial discussions, a prototype challenge\cite{TrackMLRamp2017} was organised during the Connecting The Dots workshop\footnote{\url{https://ctdwit2017.lal.in2p3.fr}} (an annual workshop for experts in pattern recognition) held at IJCLab in Orsay in March 2017. 
The problem was essentially the same as the one exposed here but significantly simplified to be a 2D problem with just 20 tracks per event (instead of 10.000 in 3D). There was no speed constraint. The same accuracy score formula was used for the first time. This 2D challenge has already yielded a variety of algorithms (not applicable in 3D though) and demonstrated that the accuracy score was indeed selecting the best algorithms.

The first Accuracy phase of the TrackML challenge proper was run on Kaggle\cite{TrackMLAccuracy2019}. It identified a variety of 3D algorithms, and a thorough investigation has shown that the Accuracy score was indeed selecting the best algorithms when their performance was evaluated using standard metrics. 

The second, Throughput, phase had significantly lower participation but it yielded a few very high quality and very fast algorithms. It is not currently possible to compare directly to in-house algorithms which would need to be adapted to this specific dataset. Also, in-house algorithms in common use usually ignore the numerous tracks with $p_T$ less than 400~MeV (the tracks with the highest curvature) while algorithms presented here are able to reconstruct tracks down to 150~MeV. So it can be estimated that in-house algorithms are at most of order 10~s per event on one CPU core, so one order of magnitude slower than Mikado from Sergey Gorbunov (a.k.a \texttt{sgorbuno}), 0.5s on two CPU cores. On the other hand, the dataset was significantly simplified (in particular neglecting sharing of points between tracks) so that it remains to be seen whether the new algorithms can live up to  expectations when used in the full ATLAS and CMS experiment context. The community is now in the process of doing this exercise.

In the end, what role can be expected for Machine Learning in tracking in the light of the TrackML challenges ? 
It does not appear that a clustering algorithm can find the track directly (as was done with DBScan based algorithms in the Accuracy phase,  which are much too slow). Of course, the field of machine learning is growing so rapidly that new algorithms might appear which would change this statement.

Nevertheless, after extended discussions between the three winners and experts in the field, a consensus appears that there are two likely avenues for the use of Machine Learning in such problems (i) combine ML with discrete optimisation, for example using a classifier to select early and quickly the best seed candidates as done by Marcel Kunze a.k.a \texttt{cloudkitchen}  (with a simple dense NN, but Graph NN seem promising) (ii) use ML to automatise the lengthy tuning of the internal parameters of the algorithms on the training dataset (circa 10.000 in the case of Mikado by Sergey Gorbunov a.k.a \texttt{sgorbuno}). 

Separately, the availability of the TrackML datasets (\cite{salzburger_andreas_2018_4730167} for the Accuracy phase and \cite{salzburger_andreas_2018_4730157} for this Throughput phase) has been extremely useful to facilitate the collaboration of experts which are usually working within their own experimental team. 
It is being used for further studies like track seeds finding with similarity hashing~\cite{Amrouche:2019yxv} or classification with deep learning~\cite{Dietrich:2019qif}, investigating the use of cluster shape to help seeding~\cite{Fox:2020hfm}, investigating tracking with graph networks~\cite{Ju:2020xty,Choma:2020cry,Duarte:2020ngm,Pata:2021oez,dezoort2021charged} (including with FPGA~\cite{Heintz:2020soy} ), investigating tracking with simulated annealing on a D-Wave quantum computer\cite{Bapst:2019llh,Zlokapa:2019tkn} or Quantum Edge Network~\cite{Tuysuz:2020ocw,Tuysuz:2020eaa,Tuysuz:2020gjh}, and building a complete generic tracking pipeline~\cite{ju2021physics}.

\section*{Acknowledgements}
The team would like to thank CERN for allowing the use of the dataset, and Codalab for hosting the competition. We are very grateful to our generous sponsors without which the challenges would not have been possible. Platinum sponsors: Kaggle, Nvidia and Universit\'e de Gen\`eve. Gold sponsors: Chalearn and DataIA. Silver sponsors : CERN Openlab, Paris-Saclay CDS, INRIA, ERC mPP, ERC RECEPT, Common Ground, Universit\'e Paris Sud/Paris Saclay, INQNET, Fermilab and pyTorch. TG acknowledges the support of the Swiss National Science Foundation under the grant 200020$\_$181984. SG acknowledges the support of the German BMBF ministry. This project has received funding from the European Union Horizon 2020 research and innovation programme under grant agreement No 724777 ``RECEPT'', No 772369 ``mPP'' and No 654168 ``AIDA-2020''. This work was made possible by Institut Pascal at Université Paris-Saclay  with the support of the program “Investissements d’avenir” ANR-11-IDEX-0003-01. In addition, the organisers would like to thank the members of the International Advisory Committee : Markus Elsing (CERN), Frank Gaede (DESY), Alison Lowndes (Nvidia), Maurizio Pierini (CERN), Danilo Rezende (Google DeepMind), Marc Schoenauer (INRIA-Saclay) and Svyatoslav Voloshynovskyy (U Genève).  

\bibliographystyle{tepml} 
\bibliography{trackml}

\end{document}